\documentclass[lettersize,journal]{IEEEtran}
\usepackage{amsmath,amsfonts}
\usepackage{algorithmic}
\usepackage{algorithm}
\usepackage{array}
\usepackage[caption=false,font=normalsize,labelfont=sf,textfont=sf]{subfig}
\usepackage{textcomp}
\usepackage{stfloats}
\usepackage{url}
\usepackage{verbatim}
\usepackage{graphicx}
\usepackage{cite}
\usepackage{multirow}
\usepackage{booktabs}
\usepackage{utfsym}
\usepackage{amsmath}
\usepackage{mathbbol}
\usepackage{pifont}
\usepackage{tikz}
\usetikzlibrary{shapes.misc}
\usetikzlibrary{trees}

\newcommand{\ie}{\textit{i.e.},~}

\usepackage[colorlinks=true,
            linkcolor=black,
            anchorcolor=green,
            citecolor=blue,
            urlcolor=orange]{hyperref}
\usepackage{colortbl}

\begin{document}

\title{Omni Survey for Multimodality Analysis \\ in Visual Object Tracking}

\author{Zhangyong Tang,
        Tianyang Xu~\IEEEmembership{Member,~IEEE,}
        Xuefeng Zhu,
        Hui Li,
        Shaochuan Zhao,
        Tao Zhou,
        Chunyang Cheng,
        Xiao-Jun Wu*,
        and Josef Kittler~\IEEEmembership{Life Member,~IEEE}
        \thanks{
        Z. Tang, T. Xu, X. Zhu, H. Li, C. Cheng,and X.-J. Wu are with the School of Artificial Intelligence and Computer Science, Jiangnan University, Wuxi 214122, China. (First author: Z. Tang: zhangyong\_tang\_jnu@163.com, Corresponding author: X.-J. Wu, e-mail: wu\_xiaojun@jiangnan.edu.cn)
        }
        \thanks{
        S. Zhao is with the School of Computer Science and Technology, China University of Mining and Technology, Xuzhou 221116, China. (e-mail: shaochuan\_zhao@cumt.edu.cn)
        }
        \thanks{
        T. Zhou is with the School of Computer Science and Engineering, Nanjing University of Science and Technology, Nanjing, 210094, China.(e-mail: : taozhou.ai@gmail.com)
        }
        \thanks{Josef Kittler is with the Centre for Vision, Speech and Signal Processing, University of Surrey, Guildford, GU2 7XH, U.K. (e-mail: j.kittler@surrey.ac.uk)}
}

\markboth{Journal of \LaTeX\ Class Files,~Vol.~14, No.~8, August~2021}%
{Shell \MakeLowercase{\textit{et al.}}: A Sample Article Using IEEEtran.cls for IEEE Journals}


\maketitle

\begin{abstract}


The development of smart cities has led to the generation of massive amounts of multi-modal data in the context of a range of tasks that enable a comprehensive monitoring of the smart city infrastructure and services.
This paper surveys one of the most critical tasks, multi-modal visual object tracking (MMVOT), from the perspective of multimodality analysis. 
Generally, MMVOT differs from single-modal tracking in four key aspects, \ie data collection, modality alignment and annotation, model designing, and evaluation. 
Accordingly, 
we begin with an introduction to the relevant data modalities, laying the groundwork for their integration.
This naturally leads to a discussion of  challenges of multi-modal data collection, alignment, and annotation.
Subsequently, existing MMVOT methods are categorised, based on different ways to deal with visible (RGB) and X modalities: programming the auxiliary X branch with replicated or non-replicated experimental configurations from the RGB branch.
Here X can be thermal infrared (T), depth (D), event (E), near infrared (NIR), language (L), or sonar (S).
The final part of the paper addresses evaluation and benchmarking.
In summary, we undertake an omni survey of all aspects of  multi-modal visual object tracking (VOT), covering six MMVOT tasks and featuring 338 references in total.
In addition, we discuss the fundamental rhetorical question: Is multi-modal tracking always guaranteed to provide a superior solution to unimodal tracking with the help of information fusion, and if not, in what circumstances its application is beneficial. 
Furthermore, for the first time in this field, we analyse the distributions of the object categories in the existing MMVOT datasets, revealing their pronounced long-tail nature and a noticeable lack of animal categories when compared with RGB datasets.
We conclude the discussions by highlighting promising directions for future research and provide a curated collection of up-to-date GitHub repositories for the MMVOT community with the link: \textcolor{blue}{https://github.com/Zhangyong-Tang/Awesome-MultiModal-Visual-Object-Tracking}.
\end{abstract}


\section{Introduction}\label{sec:introduction}
\begin{figure*}[t]
\centering
\includegraphics[width=1.0\linewidth]{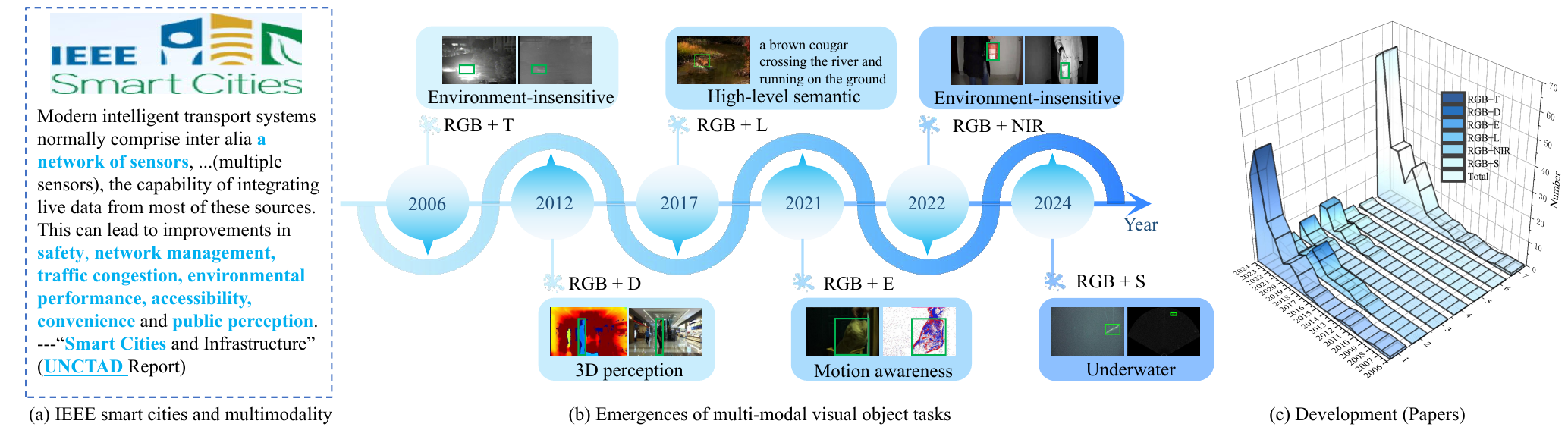}
\caption{(a) Multi-modal sensing is an essential element of smart cities. (b) The history of multi-modal visual object tracking. (c) The development of all MMVOT tasks in terms of papers published during 2006 to 2024.}
\label{fig:timeline}
\end{figure*}

\IEEEPARstart{F}{acilitating} the development of smart cities is a forward-looking yet challenging objective in the field of artificial intelligence.
In these environments, intelligent and autonomous agents are expected to handle complex decision-making processes while ensuring public safety.
However, this goal remains difficult to achieve due to the dynamic and unpredictable nature of real-world scenarios, particularly when agents operate with incomplete or fragmented perceptions of their surroundings.
To address this, as shown in Fig.~\ref{fig:timeline}(a), researchers are increasingly focused on aggregating multi-modal information to construct a more reliable understanding of the environment, thereby enhancing the robustness of agents in all-day, all-weather applications \cite{feifeili-multimodal, lrrnet}.

\begin{table*}[t]
    \centering
        \caption{A differentiation from the existing surveys and reports. `Ref. Num' means the number of references.
        }
    \footnotesize
    \resizebox{1\linewidth}{!}{
    \begin{tabular}{|ccccccc|}
    \hline
    \rowcolor[gray]{0.9}
      Type & Task &  Citation & Year & Venue  & Ref. Num&Involvement \\
      \hline \hline
      Survey & RGB+T (2)& \cite{rgbt-zhang2020} & 2020 & INFFUS & 142&   Methods; Benchmarks; Future works \\
      Survey &  RGB+T (2)& \cite{rgbt-tang2022} & 2022 & Arxiv& 56 & Methods; Benchmarks; Future works \\
      Survey &  RGB+T (2)& \cite{rgbt-zhang2023} & 2023 & TOMM& 114 & Methods; Benchmarks; Future works\\
      Survey &  RGB+T (2)& \cite{rgbt-zhang2024} & 2024 & TIM& 158 & Methods; Benchmarks; Future works\\
      Survey &  RGB+T (2)& \cite{rgbt-feng2024} & 2024 & INFFUS & 168 & Methods; Benchmarks; Future works\\
      Survey &  RGB+D (2)& \cite{rgbd-yang2022} & 2022 & Arxiv& 75 & Methods; Benchmarks; Future works\\
      Survey &  RGB+D (2)& \cite{rgbd-ou2024} & 2024 & CADCG& 86 & Methods; Benchmarks; Future works\\
      Survey &  RGB+T/D (3)& \cite{rgbdt-zhang2024} &2024& CVM & 127 & Methods; Benchmarks; Future works\\
      Report &  RGB+T/D/E/L (5)& \cite{rgbdtel-zhang2024} &2024& Arxiv & 154 & Methods; Benchmarks \\
      \hline
      & & && & &  \textbf{Multi-Modal Data Collection\&Annotation\&Alignment;}\\
            \multirow{-2}{*}{Survey} & \multirow{-2}{*}{\textbf{RGB+T/D/E/L/S/NIR (6)}} & \multirow{-2}{*}{Ours} & \multirow{-2}{*}{-}& \multirow{-2}{*}{-} & \multirow{-2}{*}{\textbf{338}} &\textbf{Methods; Benchmarks; Future works}\\
      \hline
    \end{tabular}
    }
    \label{tab:survey}
\end{table*}

\begin{figure}[t]
\centering
\includegraphics[width=1.0\linewidth]{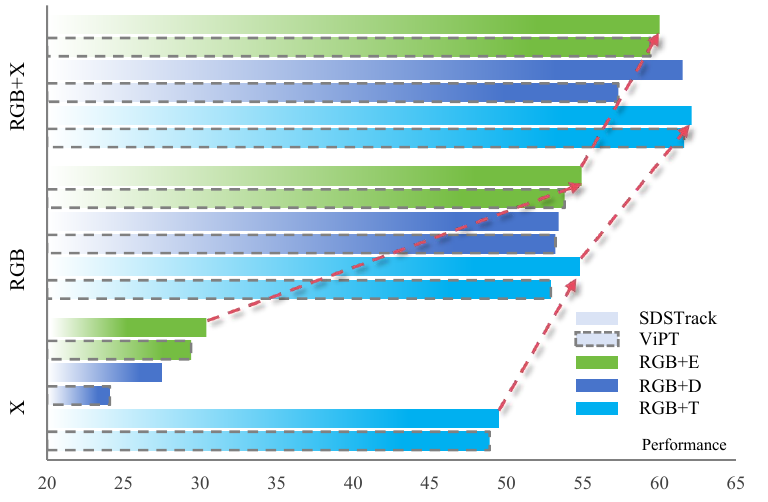}
\caption{An illustration of the evolution of the use of multi-modal data through experimental studies involving RGB+T, RGB+D, and RGB+E tracking tasks.}
\label{fig:results}
\end{figure}

As one of the most critical competences enabling motion perception in intelligent systems, such as embodied robotics \cite{robotics} and autonomous vehicles \cite{wu2023motion}, visual object tracking (VOT) draws increasing attention and remains a prominent area of research in computer vision \cite{chen2021hybrid, vot2023, siamfc, ostrack}.
The goal of VOT is to maintain awareness (location and size) of an object, given its position in the first frame of an RGB sequence \cite{siamfc}.
Analogously, multi-modal visual object tracking (MMVOT) tasks pursue the same objective but rely on multiple input signals (typically two streams), as illustrated in Fig.\ref{fig:timeline}(b), to provide more robust perception in challenging conditions with the results exemplified in Fig.~\ref{fig:results}.
In recent years, driven by the widespread availability of sensors that capture complementary information to RGB data, interest in MMVOT has surged, as 
clearly shown in Fig.\ref{fig:timeline}(c).

The aim of this work is to reflect on this rapid development and to present a comprehensive survey of MMVOT tasks from the perspective focusing on multimodality.  
\textit{Beyond the taxonomy of existing methods, it provides in-depth analyses of the physical principles underlying various data modalities, strategies for aligning and annotating multi-modal data, as well as the evaluation metrics and datasets used for performance assessment—thus offering a holistic overview of multimodality in MMVOT.}
Furthermore, this work systematically investigates two significant, yet under-explored issues faced by the MMVOT community: (1) the relevance of applying multi-modal information fusion densely and indiscriminately in all scenarios, and (2) the scarcity of certain type of data, e.g. animals, in non-RGB streams, and the much more pronounced long-tail distribution of object categories compared with that in the RGB datasets. These issues are identified and discussed in this paper for the first time in this field.
Finally, the paper concludes by outlining promising directions for future research.
All relevant materials are made publicly available at our  \href{https://github.com/Zhangyong-Tang/Awesome-MultiModal-Visual-Object-Tracking}{homepage}, which will be continuously updated to further benefit this community.

\textbf{Related Surveys and Report:} Given the rapid growth of interest in MMVOT tasks, we believe that an up-to-date and comprehensive survey from a holistic perspective is of vital importance to the community.
Table.~\ref{tab:survey} lists the existing surveys \cite{rgbt-zhang2020, rgbt-tang2022, rgbt-zhang2023, rgbt-zhang2024, rgbt-feng2024, rgbd-yang2022, rgbd-ou2024, rgbdt-zhang2024} and a report \cite{rgbdtel-zhang2024}.
Most existing surveys focus on one or two MMVOT tasks, such as RGB+T, RGB+D, or RGB+D/T tracking.
While the  report in \cite{rgbdtel-zhang2024} covers a relatively wider range of tasks, it lacks a comprehensive and in-depth analysis.
This limitation motivates our work to encompass a broader spectrum of MMVOT tasks and provide detailed discussions and insights to further accelerate progress in the field.

\textbf{Contributions:} The contributions of  our review of the landscape of the MMVOT developments can be summarised as follows:

$\bullet$ This work presents the first comprehensive survey encompassing RGB+T, RGB+D, RGB+E, RGB+L, RGB+NIR, and RGB+S tracking tasks. With a total number of 338 references, it offers readers a concise yet thorough understanding of multi-modal visual object tracking.

$\bullet$ Our survey of multi-modal VOT approaches is holistic, covering all aspects of MMVOT tasks, ranging from multi-modal data collection to multi-modal method evaluation.

$\bullet$ This work systematically discusses two significant yet previously overlooked issues: \ding{172} The need for indiscriminate and dense application of multi-modal fusion across all scenarios; \ding{173} The biased nature of object category distributions in MMVOT datasets, which exhibit clear long-tail characteristics that may hinder generalisation. Additionally, this work reveals, for the first time in the MMVOT community, the scarcity of animal data in non-RGB video streams.

$\bullet$ We also outline the warranted and promising directions for future MMVOT research, 
such as physics-inspired architectures and algorithms for modality customisation and developing metrics to quantify the quality of the respective modalities available for information fusion, aiming to uncover the hidden relationships between information fusion and tracking performance.

\textbf{Scope:} Before proceeding with the detailed analysis, we clarify the scope of this work:  (1) The focus is on single object tracking - each sequence contains only one target trajectory; for multi-object tracking, please refer to \cite{mot}; (2) Generic object tracking - category-specific tracking methods, such as pedestrian tracking \cite{pedestraintracking}, are not considered; (3) 2-D object tracking - this survey focuses on bounding box-based tracking. For 3D object tracking, readers are referred to \cite{3dtracking}.

\begin{figure*}[t]
\centering
\includegraphics[width=1.0\linewidth]{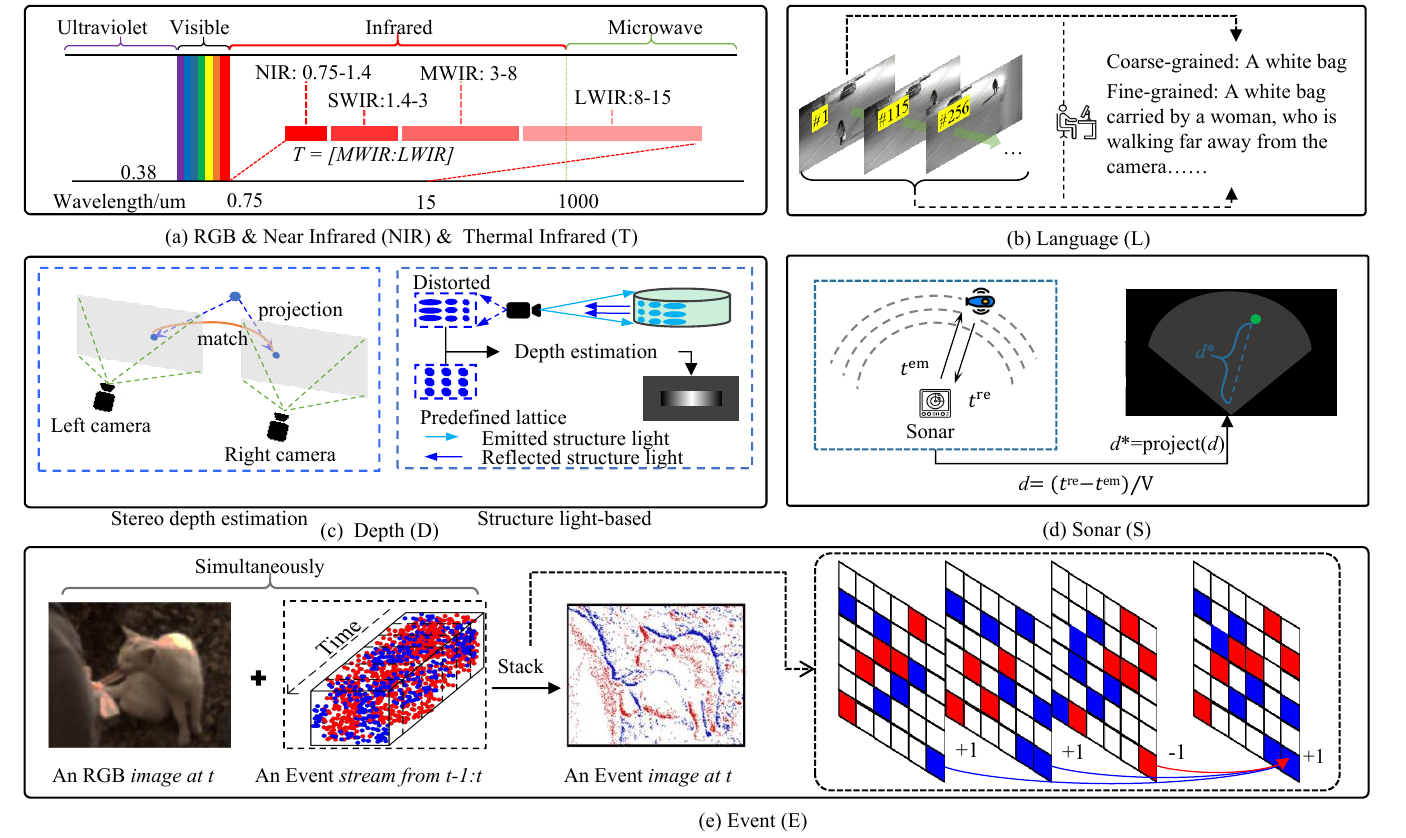}
\caption{Illustrations of the involved data modalities, including (a) RGB, T, NIR, (b) L, (c) D, (d) S, and (e) E modalities.
}
\label{fig:datamodalities}
\end{figure*}

\section{Multi-Modal Data Collection}\label{sec:data}
To facilitate an in-depth understanding, first and foremost, \textbf{we begin by providing a physical introduction to the data modalities used for tracking, highlighting their unique and 
complementary characteristics in relation to RGB data, which is always ignored in related works}.

\subsection{Data Modalities}\label{sec:datamodalities}

\textit{Visible (RGB), Near Infrared (NIR), and Thermal Infrared (T)}: As depicted in Fig.~\ref{fig:datamodalities}(a), RGB data is captured by sensing the reflected visible spectrum with a wavelength between 0.38 and 0.75 um\footnote{https://en.wikipedia.org/wiki/Visible\_spectrum}.
Electromagnetic waves at different wavelengths exhibit distinct properties.
For example, RGB data inherits the characteristics of the visible spectrum, making its imaging process (closely related to its quality) sensitive to environmental factors.
This motivates the incorporation of other electromagnetic waves.
Infrared radiation, emitted by objects themselves and less influenced by external conditions, typically begins at wavelengths longer than the red portion of the visible spectrum (0.75–1000 µm\footnote{https://en.wikipedia.org/wiki/Infrared}).
The infrared spectrum is further categorised into five parts: Near-Infrared (NIR, 0.75–1.4 µm), Short-Wavelength Infrared (SWIR, 1.4–3 µm), Mid-Wavelength Infrared (MWIR, 3–8 µm), Long-Wavelength Infrared (LWIR, 8–15 µm), and Far-Infrared (FIR, 15–1000 µm). 
Among these, MWIR and LWIR are collectively referred to as thermal infrared (T), which is particularly valuable for detecting thermal radiation emitted by objects themselves, thus being robust to environmental conditions.

\textit{Language (L)}: As shown in Fig.~\ref{fig:datamodalities}(b), unlike the conventional bounding box-based annotations, language offers a more natural and intuitive way for humans to describe objects, providing richer high-level semantic information \cite{tnl2k, vasttrack}.

\textit{Sonar (S)}: In Fig.~\ref{fig:datamodalities}(d), sonar sensors operate by actively emitting ultrasound waves at a specific timestamp $t^{\rm em}$. 
When these waves encounter an object, they are reflected back to the sensor and received at a later timestamp $t^{\rm re}$.
The time difference between emission and reception, $\Delta t = t^{\rm re} - t^{\rm em}$, is used to calculate the distance $d$ between the sonar sensor and the object, given the known speed of sound, $\rm V$, as follows: $d = (\Delta t * \rm V)/2$.
Finally, according to predefined rules, this distance is projected into the image space to create the sonar image: $d^* = \rm project(\textit{d})$, which depicts 
distance information.

\textit{Depth (D)}: Depth information can be acquired through three main approaches: Time-of-Flight (ToF) methods, stereo depth estimation, and structured light-based techniques \cite{d2cube}. 
The principle of ToF-based methods is similar to that of sonar data, relying on the temporal gap between emitting and receiving signals.
For this reason, ToF is not illustrated in Fig.~\ref{fig:datamodalities}(c), while the other two methods are depicted.
On the left side, stereo depth estimation uses two cameras that mimic the human visual system by capturing two slightly different views of the same scene.
By comparing the disparity between these views, along with camera-specific parameters such as viewpoint and baseline distance between the cameras, the depth of objects can be estimated accurately.
The right side of Fig.~\ref{fig:datamodalities}(c) illustrates the structure light-based method.
In this approach, a camera projects a known light pattern (such as a grid or lattice) onto the object. 
The deformation of the pattern caused by the surface of an object is captured in the reflected light.
By analysing this distortion, combined with internal camera parameters, the depth information of the object is derived.
Additionally, stereo depth estimation is generally recommended for outdoor scenarios due to its robustness in such environments \cite{d2cube}.

\textit{Event (E)}: 
Event cameras are biologically inspired by neurons, with values representing activation or inactivation.
Unlike regular RGB cameras that capture intensity images in a frame-based manner with relatively high latency (10–20 ms), event cameras output a stream of asynchronous events.
They record discrete changes in pixel intensity, indicated by positive (+1) or negative (–1) values, depending on whether the pixel intensity increases or decreases, respectively.
This allows event cameras to perceive rapid motions with very low latency (1 us) \cite{visevent}.
As illustrated in Fig.~\ref{fig:datamodalities}(e), 
during the acquisition of the $t$-th RGB frame, an event stream spanning from $t-1$ to $t$ is obtained.
This stream is considered the raw event data and can be represented as a voxel grid \cite{fe108}.
More commonly, the data is stacked to form the $t$-th event image \cite{visevent, crsot} by aggregating values at the same spatial location across all event slices, as shown in the right side of Fig.\ref{fig:datamodalities}(e).

\subsection{Data Collection} \label{sec:datacollection} 
After identifying the physical characteristics of each data modality, their complementary nature becomes apparent, encouraging researchers to explore multi-modal integration to develop more robust intelligent systems \cite{clip, feifeili-multimodal}.
Among the modalities discussed in this work, the L modality is unique in that it is derived from image data, with its description provided manually in early datasets \cite{lasot}, and more recently generated using large language models (LLMs) \cite{mgit}.
In contrast, other modalities are acquired through dedicated sensors.
Typically, two cameras are mounted on the same platform to capture different modalities, such as the DLD-J18-161 for T and the DS-2ZCN3007 for RGB in \cite{lasher}, or an HD Zoom SeaCam underwater camera alongside an Oculus MD-Series sonar in \cite{rgbs50}. 
While hand-crafted platforms integrating multiple sensors are widely used for data collection, they often demand a significant effort to ensure precise multi-modal alignment \cite{vtuav}. 
Alternatively, off-the-shelf systems such as the ZED 2i stereo camera\footnote{https://www.stereolabs.com/en-cn/store/products/zed-2i} simplify the process by providing factory-calibrated setups, thereby eliminating the need for complex sensor calibration.
However, these commercial solutions can be prohibitively expensive\footnote{https://store.dji.com/cn/product/zenmuse-h30t-and-dji-care-enterprise-basic?from=site-nav\&vid=167271}, making hand-crafted platforms a more practical choice for many researchers in the community.

Typically, MMVOT tasks involve two separate data streams, with the exception of the RGB+NIR task.
In this case, RGB and NIR data coexist within the same sequence. 
This is made possible by specialised cameras designed for RGB+NIR tracking, which can adaptively switch between RGB and NIR sensors in response to varying illumination conditions \cite{cmotb-aaai2022}.

\section{Multi-Modal Data Alignment and Annotation}\label{sec:anno}
\subsection{Data Alignment}
Unlike single-modal data, multi-modal data typically requires alignment to address spatial discrepancies caused by different sensor placements.
However, within the scope of this work, the RGB+NIR, RGB+L, and RGB+S datasets do not require such alignment, for the following reasons:
RGB+NIR tracking involves a single video stream containing both modalities, captured by cameras that switch adaptively between RGB and NIR modes, making the data alignment-free \cite{cmotb-tnnls2024}.
RGB+L data is semantically aligned, as the linguistic descriptions are generated from RGB images. 
Moreover, the differing data formats (text vs. image) render traditional spatial alignment methods inapplicable \cite{tnl2k}.
RGB+S data, although both in image format, differs fundamentally in content. 
RGB captures visual appearance, while S encodes distance information.
This semantic mismatch precludes meaningful spatial alignment \cite{rgbs50}.

In contrast, alignment is essential for the three most widely studied MMVOT tasks: RGB+T, RGB+D, and RGB+E tracking.
For RGB+T datasets, the most common alignment strategies involve computing transformation matrices using either intrinsic and extrinsic camera parameters \cite{gtot, vtuav, lasher, rgbt210, rgbt234} or manually annotated key points \cite{mv-rgbt, otcbvs, litiv}.
For RGB+D datasets, alignment is typically handled internally by the hardware. Examples include stereo cameras like the ZED 2i \cite{rgbd1k}, smartphones such as the iPhone \cite{arkittrack}, and depth sensors like the Kinect \cite{ptb, d2cube, depthtrack, cdtb} and ASUS Xtion \cite{stc}.
RGB+E datasets are generally aligned within the sensor hardware, such as Dynamic Vision Sensors (DVS) \cite{dvs, visevent}. Notably, CRSOT \cite{crsot} is an exception, as it is collected using an RGB camera and a higher-resolution event sensor without alignment, aiming to explore fusion strategies under unaligned conditions.

\subsection{Data Annotation}
Among the discussed modalities, L is unique in that it inherently serves as a form of high-level semantic annotation and therefore does not require traditional labelling.
In contrast, all other modalities (RGB, T, D, E, NIR, S) rely on conventional annotation methods, typically involving the use of bounding boxes to identify target objects.
Generally, two standard formats are employed to represent bounding boxes: [$x_{\rm tl}$, $y_{\rm tl}$, $x_{\rm br}$, $y_{\rm br}$] \cite{gtot} and [$x_{\rm tl}$, $y_{\rm tl}$, $w$, $h$] \cite{rgbt234, lasher, visevent}, where subscripts “$\rm tl$” and “$\rm br$” denote the top-left and bottom-right corners, respectively.
$x$ and $y$ represent the coordinates on the horizontal and vertical axes;
$w$ and $h$ refer to the width and height of the bounding box.
Thus, the first format defines a box using the coordinates of two opposing corners, while the second uses the top-left corner and the box dimensions.

In general, annotation requires experts familiar with the tracking process to ensure the precision and quality of the bounding boxes.
However, this manual approach is both time-consuming and labour-intensive, which has motivated the adoption of semi-automatic annotation methods.
In such methods, automatic tools, such as LabelMe\footnote{https://github.com/wkentaro/labelme}, are first used to generate coarse bounding boxes.
These initial annotations are then manually refined to ensure accuracy.
This approach is particularly efficient for simple or less dynamic videos, where the automatically generated boxes require minimal adjustment, thereby significantly reducing the overall annotation effort.

\section{Multi-Modal Model Designing}\label{sec:modeldesigning}
\begin{figure}[t]
\centering
\includegraphics[width=1.0\linewidth]{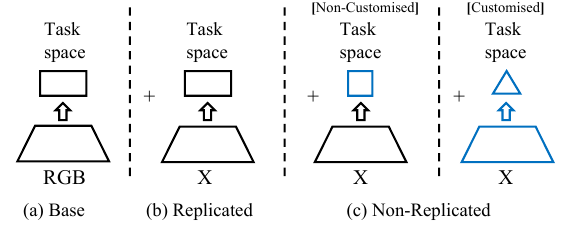}
\caption{A conceptual illustration of the taxonomy of the MMVOT methods.
}
\label{fig:taxonomy4method}
\end{figure}

As a rapidly emerging topic in the VOT community, MMVOT methods inherit many architectural designs and tracking strategies from RGB-only methods.
Let $\rm T_{RGB}: D_{RGB} \to \mathbb{R}^4$ denotes an RGB tracker, mapping the RGB input dataset $\rm D_{RGB}$ to bounding boxes in $\mathbb{R}^4$.
Analogously, a multi-modal tracker can be defined as $\rm T_{MMVOT}: D_{MMVOT} \to \mathbb{R}^4$, where $\rm D_{MMVOT}$ represents the multi-modal dataset and $\mathbb{R}^4$ remains the shared output space of the bounding boxes.
This formulation highlights the strong methodological parallels between RGB-only VOT and MMVOT tasks.
\textit{Accordingly, taking the RGB branch as a reference, existing MMVOT methods can be classified into two main categories based on the design of the additional modality branch (X branch, X $\in$ [T, D, E, NIR, L, S])}: (1) Replicated Configuration - where the X branch mirrors the architecture of the RGB branch, and (2) Non-Replicated Configuration - where the X branch adopts a different architecture.
Moreover, the methods with non-replicated configurations are specified with configurations customised or non-customised to the input of X branch.
This taxonomy is conceptually illustrated in Fig.~\ref{fig:taxonomy4method}.

\begin{figure*}[t]
\centering
\includegraphics[width=1.0\linewidth]{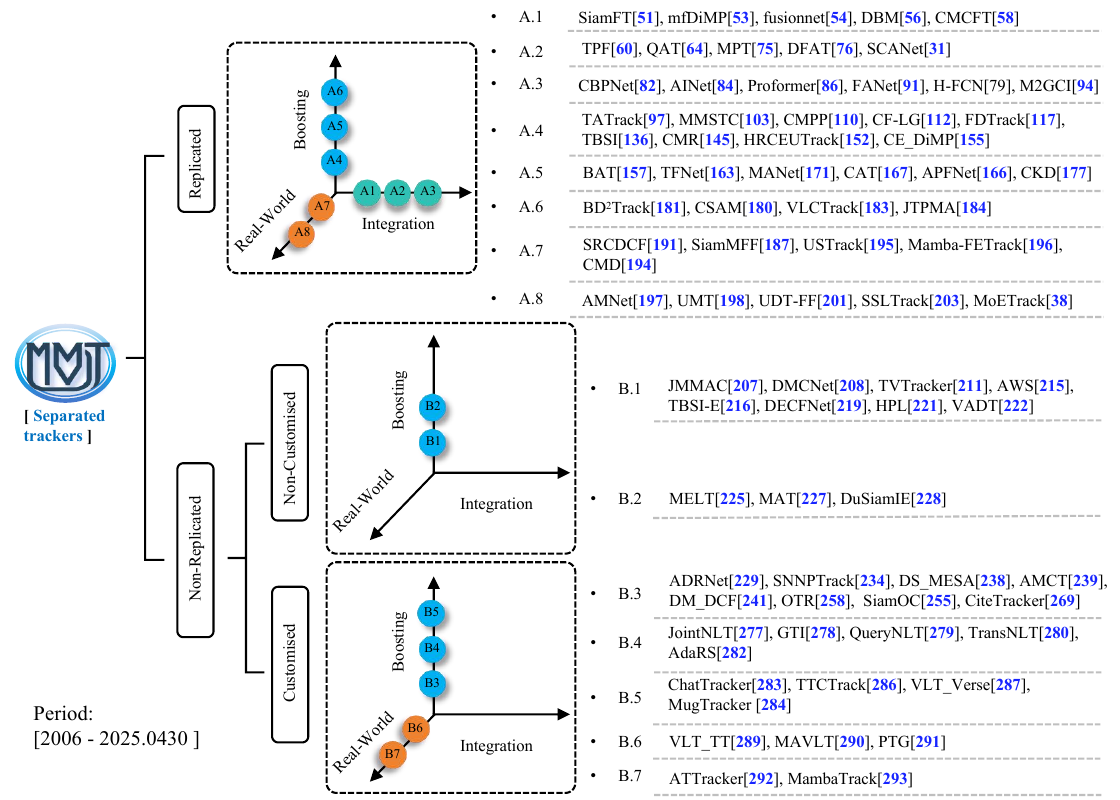}
\caption{The taxonomy of MMVOT methods according to the modality configurations and the corresponding representative trackers. A conceptual illustration of the taxonomy is provided in Fig.~\ref{fig:taxonomy4method}.
}
\label{fig:taxonomy4separated}
\end{figure*}

As shown in Fig.~\ref{fig:taxonomy4separated}, for clarity, the methods within each category are further grouped into three kinds: the methods primarily focusing on the straightforward information \textit{Integration}, the methods learning \textit{Boosted} feature representations, and the methods addressing \textit{Real-world} issues.
These sub-categories span a comprehensive design space for the MMVOT methods, offering a structured perspective for analysing and comparing different approaches.

\subsection{X Branch with Replicated Configurations}
The MMVOT methods that adopt the same experimental configurations for both the RGB and X branches fall into this category.

\textbf{A.1} Early Stage Information Fusion. 
In the early stages, several attempts are made to simply aggregate the advantages of RGB and X modalities with features extracted by convolutional neural networks (CNNs) \cite{rf-cff, cf-rgbd, lmdcn, siamft, wf_dimp, mfdimp, fusionnet, afcf}, deep Boltzmann machines \cite{dbm}, or hand-crafted descriptors such as histogram of oriented gradients (HOG \cite{hog}) \cite{cmcft}. 
Although these methods demonstrate improved performance over single-modality approaches, they do not take advantage of the power of deep learning for information fusion.

\textbf{A.2} Exploring Single-Level Fusion. 
Building on the rationale that the features at different network stages capture distinct semantic information \cite{SiamRPN++}, researchers have begun to investigate the benefits of fusion at different levels, including pixel-level fusion \cite{tpf, plf, dapf}, feature-level fusion \cite{rfc, qat, awcm, rgbs50, cmotb-aaai2022, cmotb-tnnls2024, siamfea, strt, rcdl, mmmpt}, and decision-level fusion \cite{dusiamrt, rmfnet, siammgt, mst, plasso-adspf, mpt, dfat, sccf}.
These investigations have collectively demonstrated the benefits drawn from each fusion level, thereby motivating further research into multi-level fusion strategies.

\textbf{A.3}  Exploring Multi-Level Fusion. 
Fusion of features from multiple levels is typically performed either simultaneously \cite{hst-rgb-d, h-fcn, hatfnet, dsiammft, cbpnet} or progressively \cite{csrdcf_rgbt, ainet, csmma, proformer, hmad, dafnet, mbafnet, siammmf, fanet, hdinet, mcssafnet}.
In M2GCI \cite{m2gci}, for instance, shallow features are fused simultaneously, while deep features are fused progressively.
Here, simultaneous fusion refers to extracting features from different levels independently and fusing them afterward, whereas progressive fusion implies that the fused embeddings from earlier stages contribute to the feature extraction in subsequent layers.
Among these, progressive fusion is more commonly adopted, as it reflects a coarse-to-fine reasoning paradigm, aligning with the hierarchical nature of visual representation \cite{lsdcf}.

However, the discriminative power of feature representations used in the aforementioned methods is not guaranteed, often resulting in suboptimal fusion.
To address this, extensive efforts have been devoted to learning more robust and informative feature embeddings, which in turn enhance the effectiveness of the fusion process.

\textbf{A.4} Learning Temporal, Frequency, and Spatial \& Channel Enhancement. 
(1) Temporal: As a video understanding task, MMVOT inherently benefits from temporal cues, which are crucial for building stable and reliable intelligent systems. 
Temporal modelling in MMVOT is commonly achieved through several strategies: maintaining online templates \cite{btmtrack, tatrack, cfbt, tabbtrack}, defining dynamic memory pools for temporal message propagation \cite{stmt, iamtrack, mfatrack, mmstc, dmstm, dtat, prototrack}, computing cross-frame attention maps \cite{mpdmt, taat, ldua-strcf, cmpp}, embedding bounding box coordinates \cite{mambavt}, and designing long-term recovery mechanisms \cite{cf-lg, lsar, oapf}.
(2) Frequency: Noting that the attention mechanism effectively functions as a low-pass filter \cite{attnlowpass}, preserving high-frequency information is of critical concern. The solutions suggested to achieve high frequency preservation include high-pass convolutional layers \cite{lminet}, Wavelet Transforms \cite{fdtrack}, max pooling operations \cite{fhat}, and Fourier Transforms \cite{feta, ffeuma}.
(3) Spatial \& Channel: Despite the advantages brought by temporal and frequency enhancements, most methods improve the representations along the spatial and channel dimensions 
by employing advanced attention mechanisms \cite{cstnet, mfenet, mtnet, sca-mma, tgtrack, ltoda, lrmwt, lcmit, gabbp, dtan, dbdfan, cmmp, cmc2r, mirnet, siamtih, tbsi, caformer}, adaptively distinguishing foreground from background \cite{rtlsr, mssr, mlsr, jsr, l1pf, tmgrmr, nrcmr, cmr, crsp, lgmg, tefnet, rsfnet, siamtih, eco-ta, emoe-tracker}, facilitating high-quality multi-modal interactions \cite{hrceutrack, migtd}, mitigating spatial discrepancies across modalities \cite{mfnet}, and swapping high-confidence channels within multi-modal embeddings \cite{cedimp}.

\textbf{A.5} Learning Upgraded Architectures and Losses.
Beside feature enhancements, modifications of the network architecture and loss functions, such as the adaption of large pre-trained models \cite{mplt, bat, ubpt}, and the construction of multi-branch architectures tailored for different modalities \cite{querytrack, tmtb, siamtfa, siamscr, tfnet, mecf} and for modality shared-specific challenging attributes \cite{dmtracker, apfnet, cat, cat++, ddfnet, fcfnet, manet, macft, iimf}, strive to achieve further  performance uplift.
Additionally, multi-modal contrastive loss functions have been employed to improve representation learning \cite{manet++, m5l, macnet, ckd, rltn}.

\textbf{A.6} Learning Facilitated by Multi-Task Cooperations.
Cooperative learning involving  multiple tasks is instrumental to gaining  further improvements by introducing auxiliary high-level guidance, in terms of multi-object tracking \cite{sgf_mdnet+rgbt, csam}, image generation \cite{bd2track}, image-text generation \cite{mkftracker, vlctrack}, and segmentation \cite{jtpma}.
Following the standard multi-task learning paradigm, the cooperative learning process can be optimised by introducing learnable weights to balance the
multi-task losses \cite{jtpma}, leading to a more harmonious and effective task integration.

Although promising performance has been achieved, several issues require further consideration for real-world deployment.

\textbf{A.7} Computational Efficiency.
Complex designs improve performance but often at the cost of lower efficiency, which remains a crucial and persistent challenge for the community.
To accelerate MMVOT methods, researchers employ high-efficiency Siamese networks \cite{dasn, siamcaf, siammff, siamivfn, siamtdr} or Discriminative Correlation Filter (DCF) frameworks \cite{bccf, srcdcf}, lightweight and one-stream backbones \cite{amrt, mplkd, cmd, ustrack} for feature extraction, as well as efficient fusion blocks such as Mamba \cite{mamba-fetrack}.

\textbf{A.8} Tracking in Adverse Scenarios.
There is also considerable interest in the robustness of tracking under adverse conditions, which is critical for real-world deployment.  This includes tracking with unaligned \cite{amnet} and unpaired \cite{umt} data, handling multi-modal incomplete scenarios \cite{mctrack, ipl, mv-rgbt}, and employing backbones trained via self-supervised learning \cite{udt-ff, s2otformer, ssltrack, ed-dcfnet}.

\subsection{X Branch with Non-Replicated Configurations}
Mainstream methods often adopt identical configurations for both RGB and X branches. This approach overlooks the inherent physical differences between modalities, resulting in underutilisation of their complementary strengths.
Motivated by this finding, many methods explore non-replicated configurations for the X branch, which can be either customised or non-customised. 
For non-customised methods, the focus is on two issues: effective leveraging of the benefits derived from the RGB modality and measure to enhance the representation of the X modality.

\textbf{B.1} Exploiting the Strength of the RGB Modality.
Compared with other data modalities, RGB data is known to excel thanks to its rich texture and colour information \cite{lasher}.
Detailed texture cues facilitate the estimation of object motion \cite{mmdff, fsbnet, jmmac} and camera movement \cite{dmcnet, jfakf}, as well as the generation of linguistic descriptions \cite{sht, tvtracker}, which provide more appearance-independent semantic information.
Colour information enables the use of Colour Name features \cite{cn} \cite{caff, pro, aws} and improves the perception of pixel intensity \cite{tbsi-e}.
To leverage both, the texture and colour information, some methods dedicate special effort to the design of the RGB branch, focusing on the shallow layers, where such cues are abundant \cite{agminet, ifenet, decfnet, cdaat}.
Another notable strength of the RGB modality is the widespread availability of visible-spectrum sensors, resulting in far larger datasets compared to other modalities.
This abundance has enabled the development of large foundation models pre-trained solely on RGB data. 
Consequently, a prominent research direction in the MMVOT field is adapting these pre-trained models to multi-modal tracking tasks through techniques such as prompt-tuning \cite{hpl}, adapter-tuning \cite{vadt, tufnet}, and style transformation \cite{afnet}, aiming to maximise the benefits of these powerful models.

\textbf{B.2} Exploiting Enhanced X Representations.
Distinct from piggybacking on the RGB modality, a distinct trend is to enhance the representations derived for the X modality by training the X branch from scratch instead of fine-tuning \cite{melt}, filtering out indiscriminate patterns through attention mechanisms \cite{fadsiamnet, mat}, and reusing X features to emphasize illumination-invariant patterns \cite{dusiamie}.

Recognising that non-customised configurations for the X branch achieve promising performance, researchers have taken a step further by designing customised X branches, tailored to the natural characteristics of each modality.
The methods in this category feature modality-specific configurations and are therefore presented in the specialist sections dedicated to different modalities.

\textbf{B.3} Building Modality-Customised Architectures.
RGB+T: Based on its inherent imaging physics principles, RGB data is sensitive to illumination changes (ILL), whereas T data is not.
Conversely, T data struggles in the presence of multiple objects with similar temperatures, which is called thermal cross-over (TC).
Inspired by these insights, researchers have designed specialist branches, targeting these attributes: an ILL branch for RGB and a TC branch for T \cite{adrnet, asfnet, saft}. Each branch is trained separately using data annotated with the corresponding attribute.

RGB+E: E data can be regarded as binary images, where only the pixel positions with detected changes are activated, making the image much sparser than other modalities and functionally similar to neurons.
Inspired by this sparsity, Pooler \cite{tenet} leverages pooling operations to process event data effectively. Additionally, the methods based on Spiking Neural Networks (SNN) \cite{snn}, such as \cite{snnptrack, cfe, siameft, mmht}, draw motivation from the biological characteristics of event data.
Furthermore, most existing methods stack continuous event slices into a single event frame, as illustrated in Fig.~\ref{fig:datamodalities}(e), which tends to  neglect fine-grained motion information.
To overcome this limitation, the event stream can be divided into multiple groups, producing several sub-frames \cite{ds-mesa} that better preserve temporal details.

RGB+D: D data augments the RGB object description by enriching it from a unique distance perspective, providing consistent or smoothly varying depth values.
Leveraging this property, depth information is used for position correction \cite{amct, seoh}, foreground/background segmentation \cite{dmdcf, depth-ccf}, and occlusion handling and indication \cite{3s-rgbd, cdg, ca3dms, dal, mcbt, tsdm, rotsl, ds-kcf, ds-kcf-shape, dohr, oapcf, fecd, siamoc, dmdcf}. They commonly exploit the consistency of the depth values in the temporal domain based on the assumption only slight changes occur between adjacent frames or determine whether there are multiple peaks of interest in depth regions \cite{rt-kcf, ohm}.  
Moreover, depth clues also provide 3-D perception, enabling the transformation of 2D images into 3D space, which allows for more precise object localisation and tracking \cite{otr, dls, 3dpt, 3d-t, rgbd-od}.

RGB+L: Compared with other modalities, the L modality is not image-based, highlighting its unique characteristics and prompting the design of specialised architectures.
Text encoders, which are tailored for processing linguistic inputs, are exclusive components of RGB+L tracking systems and do not appear in other multi-modal tracking tasks.
Beyond this distinction, the development of RGB+L tracking largely parallels that of other modalities, with considerable efforts devoted to spatio-temporal enhancement \cite{mmtrack, mambavlt, rttnld, satracker, aclip, aitrack, citetracker, mfavlt, snlt, ovlm, vlatrack, allinone, capsulenlt}.

\textbf{B.4} Learning Facilitated by  Multi-Task Cooperation.
RGB+L: Due the inherent similarity of RGB+L tracking and visual grounding tasks, aiming to localise objects based on both visual-linguistic vs. linguistic inputs, numerous methods have explored cooperative learning between these two domains \cite{uvltrack, jointnlt, gti, querynlt, transnlt, pgg}.
In additon, AdaRS \cite{adars} introduces a novel approach by treating linguistic descriptions as information queries. It retrieves candidate regions from RGB images based on high semantic similarity, thereby facilitating an effective integration of RGB and L modalities.

\textbf{B.5} Learning Refined X Representations.
RGB+L: As introduced in Sec.~\ref{sec:datamodalities}, linguistic annotations in RGB+L tracking typically describe the target object in the entire video sequence, specific video segments, or only the initial frame, rather than providing frame-level descriptions.
This coarse granularity implies that the annotations may not fully capture the dynamics of object appearance across time, and inaccuracies in the textual descriptions further exacerbate the issue.
To address these limitations, recent methods have explored refining text embeddings through updating textual inputs during tracking \cite{chattracker, mugtracker}, adaptively selecting the most discriminative linguistic patterns \cite{osdt, ttctrack}, and maintaining fine-grained attribute-level descriptions \cite{vltverse, vlt_ost}.

\textbf{B.6} Tracking with High Efficiency.
RGB+L: Efficiency remains a long-standing and critical topic in the tracking community.
Accordingly, several RGB+L tracking methods have focused primarily on improving efficiency.
Specifically, this is achieved through the adoption of lightweight backbone networks \cite{vlt_tt}, the use of more efficient fusion modules such as Mamba \cite{mavlt}, and strategies that compress input information into fewer tokens \cite{ptg}.

\textbf{B.7} Tracking in Adverse Scenarios.
RGB+L: Robustness in adverse scenarios is also of significant concern.
Hence, \cite{attracker} explores a semi-supervised learning paradigm to mitigate the challenges posed by limited annotations and \cite{mambatrack} focuses on enhancing the tracking performance in nighttime conditions, specifically within Unmanned Aerial Vehicle (UAV)-based settings.

\begin{table*}[t]
\scriptsize
\caption{An overview of existing unified MMVOT Trackers. Please find the definition of "Uni-A" and "Uni-B" in Sec.\ref{sec:method-uni}.}
\resizebox{1\linewidth}{!}{
\begin{tabular}{|c|cccccccc|c|}
\hline
    \rowcolor[gray]{0.9}
\multicolumn{1}{|c}{\multirow{1}{*}{Unification}}     &   Method & Venue &Backbone& LasHeR  & DepthTrack & VisEvent & TNL2K & RGBS50 & Descriptions                 \\ \hline \hline
\multicolumn{1}{|c|}{\multirow{14}{*}{Uni-A}} & ProTrack \cite{protrack}   & ACM MM'22 & ResNet \cite{resnet} & 42.1 & 57.8 & 47.4& -& - &Pioneering work; Prompt-tuning\\ 
& ViPT  \cite{vipt}     & CVPR'23&ViT-B& 52.5 & 59.4     & 59.2 & -    & - &Milestone; Prompt-tuning \\ 
& FDAFT \cite{fdaft}      &PRCV'23 &ViT-B \cite{vit}&55.3 & 62.0     & - & -     & - & -   \\
& MINet  \cite{minet}   & IVC'24 &ViT-B& 52.9 & 60.4     & 59.4 & -     & - & -\\
& MixRGBX \cite{mixrgbx}    & NeuroCom'24&MixFormer \cite{mixformer}&53.6 & 60.1     & 60.2 & -     & - & - \\
 &AMATrack \cite{amatrack}& TIM'24   &ViT-B& -     & 61.8 &- &- &-&Generative Model; Tested on RGBT234\\
 & KSTrack  \cite{kstrack}  & TCSVT'24 &ResNet&-     & 58.7     & 57.2 & -     & -&Modality incomplete\\
 & GMMT \cite{gmmt}     & AAAI'24  &ViT-B& 56.6 & - & -& -&-&Also tested on RGBD1K\\ 
& OneTracker \cite{onetrack}& CVPR'24 &ViT-B& 53.8 & 60.7 & 58.0&-&-&-     \\ 
& SDSTrack  \cite{sdstrack}& CVPR'24 &ViT-B& 53.1 & 61.4     & 59.7 &-&-&- \\ 
& SeqTrackV2 \cite{seqtrackv2}& Arxiv'24&ViT-L& 61.0 & 62.3     & 63.4& 62.4 &-&- \\
 & STTrack \cite{sttrack}   & AAAI'25&ViT-B& 60.3 & 63.3 & 61.9& - & -&-\\
 & CMDTrack \cite{cmdtrack} & TPAMI'25  & ViT-B & 56.6 & 59.8 & 61.3 & - & -&Efficiency; Knowledge distillation \\
 & LightFC-X \cite{lightfcx} & Arxiv'25 &ViT-T & 50.7 & 53.8 & 53.2 & -     & 58.6/39.5&Efficiency \\
 \hline
 \multicolumn{1}{|c|}{\multirow{7}{*}{Uni-B}} & EMTrack \cite{emtrack}  &TCSVT'24 &ViT-B& 53.3     &58.3 & 58.4 & -     & -&Efficiency;  Knowledge distillation\\
 &Un-Track \cite{unitrack}  &CVPR'24&ViT-B& 51.3 & 61.0     & 58.9 & -     & - & Still require task prior\\
  & XTrack \cite{xtrack}   &Arxiv'24&ViT-B & 52.5 & 59.7     & 59.1 & -     & - & Mixture-of-expert \\ 
  &M\textsuperscript{3}Track \cite{m3track}   & SPL'25&ViT-B& 51.1 & 59.7     & 59.2 & -     & -&Meta-learning \cite{metalearning} \\ 
 & APTrack \cite{aptrack}    &TAI'25 &ViT-B&58.9 & 62.1     & 61.8 & -&-&- \\
  & UASTrack \cite{uastrack}  &Arxiv'25&ViT-B& 57.0 & 62.8 & 61.0 & -&-&Modality-specific adapters    \\
 & SUTrack \cite{sutrack} & AAAI'25 &Fast\_iTPN-L \cite{fastitpn}   & 61.9 & 66.4 & 63.8 &67.9     & -&Training with more data (RGB data) \\ \hline
\end{tabular}
\label{tab:unified}
}
\end{table*}

\begin{figure}[t]
\centering
\includegraphics[width=1.0\linewidth]{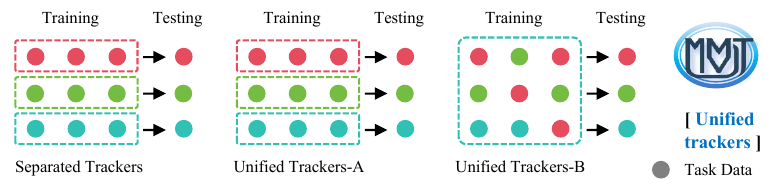}
\caption{Training and testing paradigms of existing unified trackers.}
\label{fig:taxonomy4unified}
\end{figure}

\subsection{Unified Multi-Modal Trackers}\label{sec:method-uni}
The methods shown in Fig.\ref{fig:taxonomy4separated} represent separate approaches, which are trained and evaluated on a single type of MMVOT task.
However, the growing complexity and uncertainty in smart city environments demand more versatile solutions that are applicable to more than one task, prompting the study of pairwise multi-modal tracking tasks concurrently.
This has led to the emergence of a new research direction, which focuses on the development of unified methods, capable of supporting multiple MMVOT tasks simultaneously, as illustrated in Fig.\ref{fig:taxonomy4unified}.
These unified methods can be categorised into two groups based on their training paradigms: Uni-A and Uni-B.
Methods in Uni-A adopt a training and testing pipeline similar to developing separate methods, where each task is handled independently, with solely a shared network architecture maintained and the parameters retrained.
In contrast, Uni-B methods train on a mixture of data from all the tasks jointly, resulting in unified models that generalise across modalities and tasks.
Detailed information on these unified methods, including publication source, year, backbone, performance, and brief descriptions, is provided in Table.~\ref{tab:unified}.
This research direction has gained momentum since its inception in 2022, with a significant surge in interest for Uni-B methods, all of which have been published in 2024 and 2025. 

It should be noted that all unified methods can also be classified following the same taxonomy used for separate methods, with the majority falling under the replicated and non-customised categories.
However, our proposed taxonomy for unified methods offers a more granular perspective, clearly delineating the boundary between two distinct types of unification and providing deeper insight into their design philosophies.

\subsection{Loss Function}
The MMVOT and VOT tasks share the same objective, i.e. predicting the scale and location of a specified object. Hence, MMVOT trackers adopt the same loss functions as those used in RGB-only methods\cite{xu2020accelerated}. In view of this familiarity, due to the space limitations, the details of the loss functions are relegated to the supplementary material.

\section{ Evaluation Methodology}\label{sec:evaluation}
In the following, we introduce the multi-modal datasets, evaluation metrics, the performance of existing methods, and the challenging attributes for detailed analysis.

\begin{table*}[t]
\centering
\caption{The statistics of real world multi-modal visual objecting tracking datasets.}
\resizebox{1\linewidth}{!}{
\begin{tabular}{|ccccccccc|}
\hline
    \rowcolor[gray]{0.9}
 &   &   & Video & Total & Avg.  &  &  &  \\
    \rowcolor[gray]{0.9}
\multirow{-2}{*}{Task}&\multirow{-2}{*}{Dataset}&\multirow{-2}{*}{Venue}&Num.&Frames&Frames&\multirow{-2}{*}{Aligned}&\multirow{-2}{*}{Annotation}& \multirow{-2}{*}{Description}\\
\hline \hline
\multicolumn{1}{|c}{\multirow{9}{*}{RGB+T}} & \href{http://vcipl-okstate.org/pbvs/bench/}{OTCBVS} \cite{otcbvs} & CVIU'06 & 6 & 8.5K & 1424  & \usym{1F5F8} & BB  & the first RGB+T dataset     \\
\multicolumn{1}{|c}{}                       & \href{https://www.polymtl.ca/litiv/en/codes-and-datasets}{LITIV} \cite{litiv} & CVIU'11 & 9 & 6.3K & 702 & \usym{1F5F8} & BB  & pioneering dataset          \\
\multicolumn{1}{|c}{}                       & \href{https://github.com/mmic-lcl/Datasets-and-benchmark-code?tab=readme-ov-file}{GTOT} \cite{gtot}  & TIP'16 & 50 & 7.8K & 157  & \usym{1F5F8} & BB  & the first dataset with certain scale\\
\multicolumn{1}{|c}{}                       & \href{https://github.com/mmic-lcl/Datasets-and-benchmark-code?tab=readme-ov-file}{RGBT210} \cite{rgbt210}& ACM MM'17 & 210  & 104.7K & 498 & \usym{1F5F8} & BB & larger size         \\
\multicolumn{1}{|c}{}                       & \href{https://github.com/mmic-lcl/Datasets-and-benchmark-code?tab=readme-ov-file}{RGBT234} \cite{rgbt234}& PR'19 & 234  & 116.7K & 498  & \usym{1F5F8} & BB  & RGBT210 + videos in some special scenarios         \\
\multicolumn{1}{|c}{}                       & \href{https://github.com/mmic-lcl/Datasets-and-benchmark-code?tab=readme-ov-file}{LasHeR} \cite{lasher}& TIP'21  & 1224 & 734.8K & 600 &  \usym{1F5F8} & BB  & contains the first large-scale training set           \\
\multicolumn{1}{|c}{}                       & \href{https://zhang-pengyu.github.io/DUT-VTUAV/}{VTUAV} \cite{vtuav}& CVPR'22  & 500 & 1.7M & 3329 & \usym{1F5F8} & BB+Mask  & captured by UAVs          \\
\multicolumn{1}{|c}{}                       & \href{https://github.com/Zhangyong-Tang/MultiModal-Visual-Object-tracking/tree/main}{MV-RGBT} \cite{mv-rgbt}& Arxiv'24 & 122 & 89.9K & 737  & \usym{1F5F8} & BB  & considerations on modality validity           \\
\hline
\multirow{8}{*}{RGB+D}                     & \href{http://www.iai.uni-bonn.de/~kleind/tracking/index.html}{BoBoT\_D} \cite{bobotd} & - & 5 & -  & - & - & - & unavailable, mentioned in \cite{amct}          \\
                                           & \href{https://tracking.cs.princeton.edu/index.html}{PTB} \cite{ptb}& ICCV'13 & 100 & 21.5K & 215  & \usym{1F5F8} & BB &  pioneering dataset          \\
                                           & \href{https://github.com/shine636363/RGBDtracker?tab=readme-ov-file}{STC} \cite{stc}& TCYB'17  & 36 & 9.0K & 250 & \usym{1F5F8} & BB  & outdoor videos           \\
                                           & \href{https://www.votchallenge.net/vot2019/dataset.html}{CDTB} \cite{cdtb}& ICCV'19 & 80 & 102.0K & 1274 & \usym{1F5F8} & BB  & more diverse  \\
                                           & \href{https://github.com/xiaozai/DeT}{DepthTrack} \cite{depthtrack}& ICCV'21 & 200 & 294.6K & 1473 & \usym{1F5F8} & BB  & the first training set  \\
                                           & \href{https://github.com/xuefeng-zhu5/RGBD1K}{RGBD1K} \cite{rgbd1k}& AAAI'23  & 1050 & 2.5M & 2384 & \usym{1F5F8} & BB  & the first large-scale training set           \\
                                           & \href{https://arkittrack.github.io/}{ARKittrack} \cite{arkittrack}& CVPR'23 & 300 & 229.7K & 765 & \usym{1F5F8} & BB+Mask  & collected by mobile phone           \\
                                           & \href{https://github.com/yjybuaa/RGBDAerialTracking}{D\textsuperscript{2}Cube} \cite{d2cube}& CVPR'23  & 1000  & 1.0M & 1030 & \usym{1F5F8} & BB  & captured by UAVs            \\
                                           \hline
\multirow{6}{*}{RGB+E}  & \href{https://prg.cs.umd.edu/BetterFlow.html}{EED} \cite{eed}& IROS'18  & 7 & 234.0 & 33 & \usym{2717} & BB & the first dataset  \\
                                            & \href{https://github.com/Jee-King/ICCV2021_Event_Frame_Tracking}{FE108} \cite{fe108}& ICCV'21  & 108 & 208.7K & 1932 & \usym{1F5F8} & BB & the first dataset with certain scale  \\
                                           & \href{https://github.com/wangxiao5791509/VisEvent_SOT_Benchmark?tab=readme-ov-file}{VisEvent} \cite{visevent}& TCYB'24 & 820 & 371.1K & 452 & \usym{1F5F8} & BB & the first large-scale training set            \\
                                           & \href{https://github.com/Event-AHU/COESOT}{COESOT} \cite{coesot}& Arxiv'22 & 1354 & 478.7K & 353 & \usym{1F5F8} & BB & maximum number of videos           \\
                                            & \href{https://zhangjiqing.com/publication/frame-event-alignment-and-fusion-network-for-high-frame-rate-tracking/}{FE240hz} \cite{afnet}& CVPR'23  & 110 & 143.0K & 1300 & \usym{1F5F8} & BB & an extension of FE108  \\
                                           & \href{https://github.com/Event-AHU/FELT_SOT_Benchmark}{FELT} \cite{felt}& Arxiv'24 & 742 & 1.6M & 2148 & \usym{1F5F8} & BB & long-term tracking      \\
                                           & \href{https://github.com/Event-AHU/Cross_Resolution_SOT}{CRSOT} \cite{crsot}& Arxiv'24 & 1030 & 305.0K & 296 & \usym{2717} & BB & unaligned multi-modal data          \\
                                           & \href{https://zhangjiqing.com/dataset/}{FE141} \cite{fe141}& IJCV'24 & 141 & 25.1K & 178 & \usym{1F5F8} & BB & an extension of FE108      \\  
                                           \hline
\multirow{7}{*}{RGB+L}                     & \href{https://github.com/QUVA-Lab/lang-tracker}{OTB99-L} \cite{otb99_l}& CVPR'17  & 100 & 59.0K& 590 & - & BB & the first dataset         \\
                                           & \href{https://github.com/HengLan/LaSOT_Evaluation_Toolkit}{LaSOT} \cite{lasot}& CVPR'19 & 1400 & 3.5M & 2506 & - & BB  & the first large-scale training set         \\
                                           & \href{https://github.com/HengLan/LaSOT_Evaluation_Toolkit}{LaSOT\_EXT} \cite{lasot_ext}& IJCV'21  & 1550 & 3.9M & 2502 & - & BB & an extension of LaSOT         \\
                                           & \href{https://github.com/wangxiao5791509/TNL2K_evaluation_toolkit}{TNL2K} \cite{tnl2k}&CVPR'21 & 2000 & 1.2M & 622 & - & BB &  linguistic descriptions for the whole video       \\
                                           & \href{http://videocube.aitestunion.com/}{MGIT} \cite{mgit}& NIPS'23 & 150 & 2.0M & 13333 & - & BB &  3 multi-granularity text descriptions for long-term videos       \\
                                           & \href{https://github.com/HengLan/VastTrack}{VastTrack} \cite{vasttrack}& NIPS'24 & 50610 & 4.2M & 83 & - & BB & the most object categories and videos     \\
                                             &\href{https://arxiv.org/abs/2409.08887}{VLT-MI} \cite{vlt-mi}& Arxiv'24 & 3619 & 6.6M & 1824 & - & BB & multiple multi-modal interactions  \\
                                           & \href{https://github.com/983632847/Awesome-Multimodal-Object-Tracking}{WebUOT-1M} \cite{webuot1m}& Arxiv'24 & 1500 & 1.1M & 733 & - & BB & underwater tracking        \\      &\href{http://videocube.aitestunion.com/downloads}{DTVLT} \cite{dtvlt}& Arxiv'24 & 13134 & 8.0M & 611 & - & BB & 4 multi-granularity text descriptions  \\
                                           \hline
\multirow{2}{*}{RGB+NIR}                                     & \href{https://github.com/mmic-lcl/Datasets-and-benchmark-code}{CMOTB-easy} \cite{cmotb-aaai2022}& AAAI'22  & 644 & 478.0K & 742  & - & BB & the first dataset         \\
                                   & \href{https://github.com/mmic-lcl/Datasets-and-benchmark-code}{CMOTB} \cite{cmotb-tnnls2024}& TNNLS'24  & 1000 & 799.0K & 799  & - & BB & an extension of CMOTB-easy         \\
\hline
RGB+S                                      & \href{https://github.com/LiYunfengLYF/RGBS50}{RGBS50} \cite{rgbs50}& TCSVT'24  & 50 & 43.7K & 874 & \usym{2717} & BB & the first dataset         \\
\hline
\multirow{3}{*}{RGB+Multi}                 & \href{https://github.com/983632847/WebUAV-3M}{WebUAV-3M} \cite{webuav3m}& TPAMI'23 & 4500 & 3.2M & 710 & \usym{1F5F8} & BB & RGB+Language+Audio         \\
                                           & \href{https://github.com/xuefeng-zhu5/UniMod1K}{UniMod1K} \cite{unimod1k}& IJCV'24  & 1050 & 2.5M & 2384 & \usym{1F5F8} & BB & RGB+Depth+Language \\
                                           & \href{https://github.com/mmic-lcl/Datasets-and-benchmark-code}{QuadTrack600} \cite{quadtrack600}& Arxiv'25  & 600 & 348.7K & 581 & \usym{1F5F8} & BB & RGB+Thermal++Event+Language, unpublished \\
                                           \hline
\end{tabular}
\label{tab:dataset}
}
\end{table*}

\subsection{Datasets}
Table.~\ref{tab:dataset} provides the key statistics for all MMVOT datasets, along with URLs to their homepages for easy access.
From the table, it is evident that the number of datasets has increased significantly in recent years, particularly since 2020, reflecting the rapid development of the MMVOT field.
Due to page limitations, only the most important information is included in Table.~\ref{tab:dataset}. More detailed descriptions are available in the supplementary material.

\subsection{Evaluation Metrics}\label{sec:Protocols}
To make this work self-contained, we introduce the widely used metrics, including precision rate (PR) \cite{lasher}, success rate (SR) \cite{lasher}, normalized precision rate (NPR) \cite{trackingnet}, precision (Pr) \cite{cdtb}, recall (Re) \cite{cdtb}, F-score \cite{cdtb}, and the VOT protocol \cite{vot-rgbt2020}.
Due to page limitations, only the theoretical concepts are presented here, while detailed mathematical formulations are provided in the supplementary material.

PR and NPR: PR measures the percentage of accurately tracked frames, where the Euclidean distance between the centres of the ground truth and predicted bounding boxes falls below a predefined threshold.
However, PR is sensitive to object size and image resolution, which can result in unfair evaluations for objects of different scales \cite{trackingnet}. 
To address this, NPR is introduced as a normalised version of PR, where the Euclidean distance is divided by the diagonal length of the ground truth bounding box to enable a scale-invariant evaluation.

SR: It is a metric similar to PR, measuring the percentage of the successfully tracked frames, where the Intersection over Union (IoU) between the ground truth and predicted bounding boxes exceeds a predefined threshold (commonly set to zero).

Pr, Re, and F-score: They are long-term metrics borrowed from \cite{cdtb}. Pr and Re measure the average accuracy when the predicted and ground truth bounding boxes are present, respectively. The F-score combines Pr and Re to provide a comprehensive metric for ranking the tracker performance. 

VOT protocol: Accuracy (A), robustness (R), and expected average overlap (EAO) are the metrics employed in the VOT protocol. 
Among them, A and R correspond closely to precision rate (PR) and success rate (SR), respectively, and EAO serves as a comprehensive metric that combines both A and R, playing a role similar to the F-score.

\subsection{Benchmarking Results}

\begin{figure*}[t]
\centering
\includegraphics[width=1.0\linewidth]{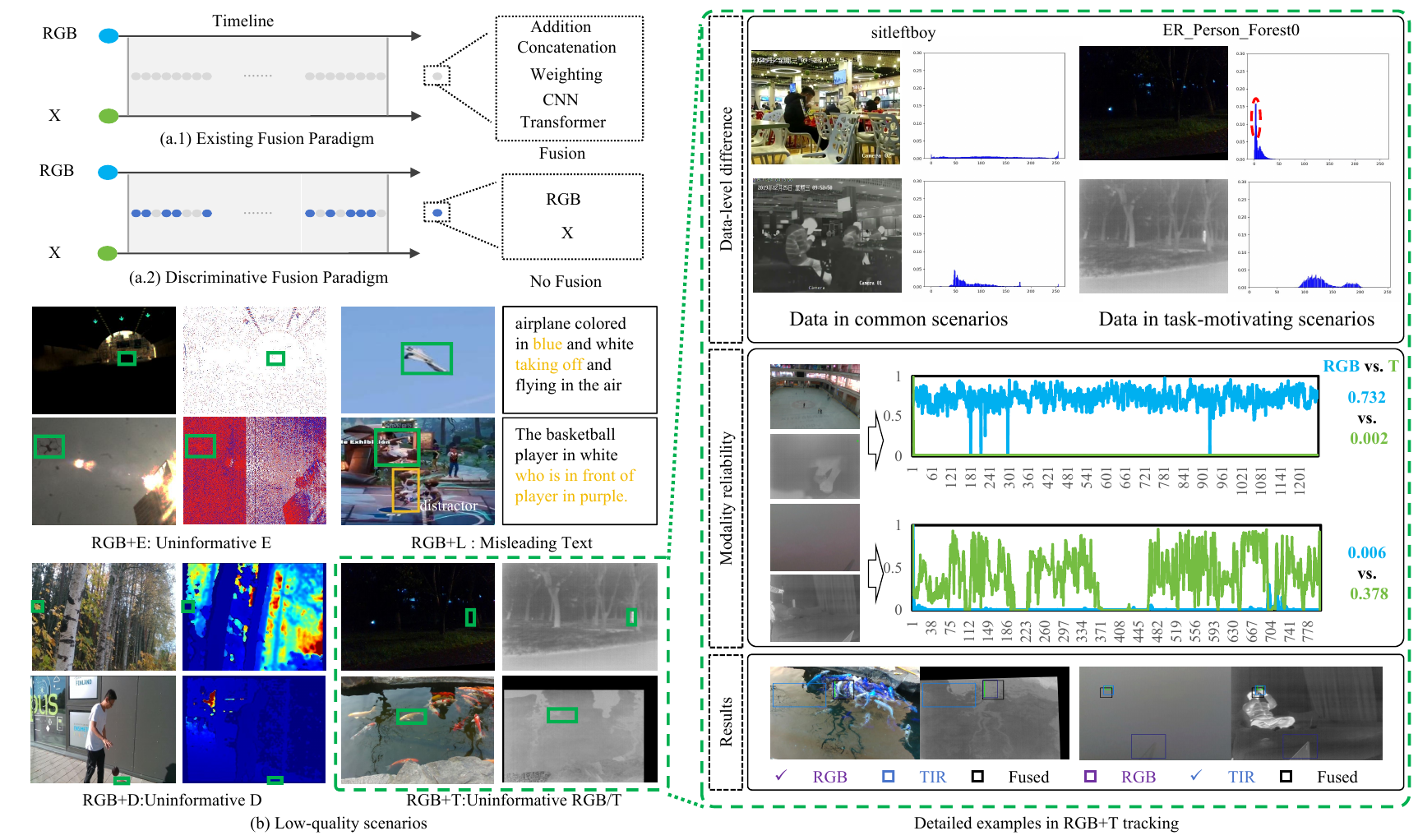}
\caption{(a) The differences between the existing and discriminative fusion paradigms. (b) An illustration of the scenarios where one modality suffers from low quality, exemplified by an example  in RGB+T tracking.
}
\label{fig:necessaryoffusion2}
\end{figure*}

\begin{figure}[t]
\centering
\includegraphics[width=0.95\linewidth]{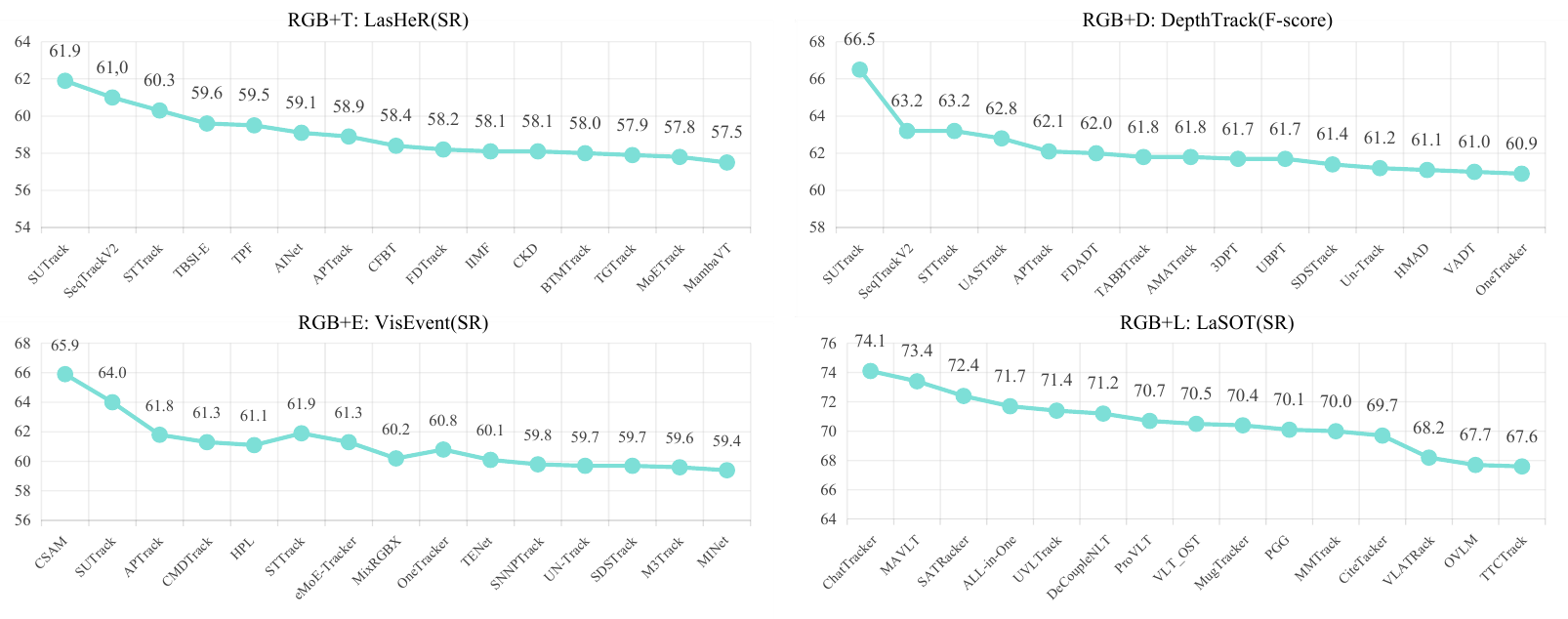}
\caption{The results on the most common benchmarks for each tracking task.
}
\label{fig:resultsonbenchmarks}
\end{figure}

\begin{figure*}[t]
\centering
\includegraphics[width=1.0\linewidth]{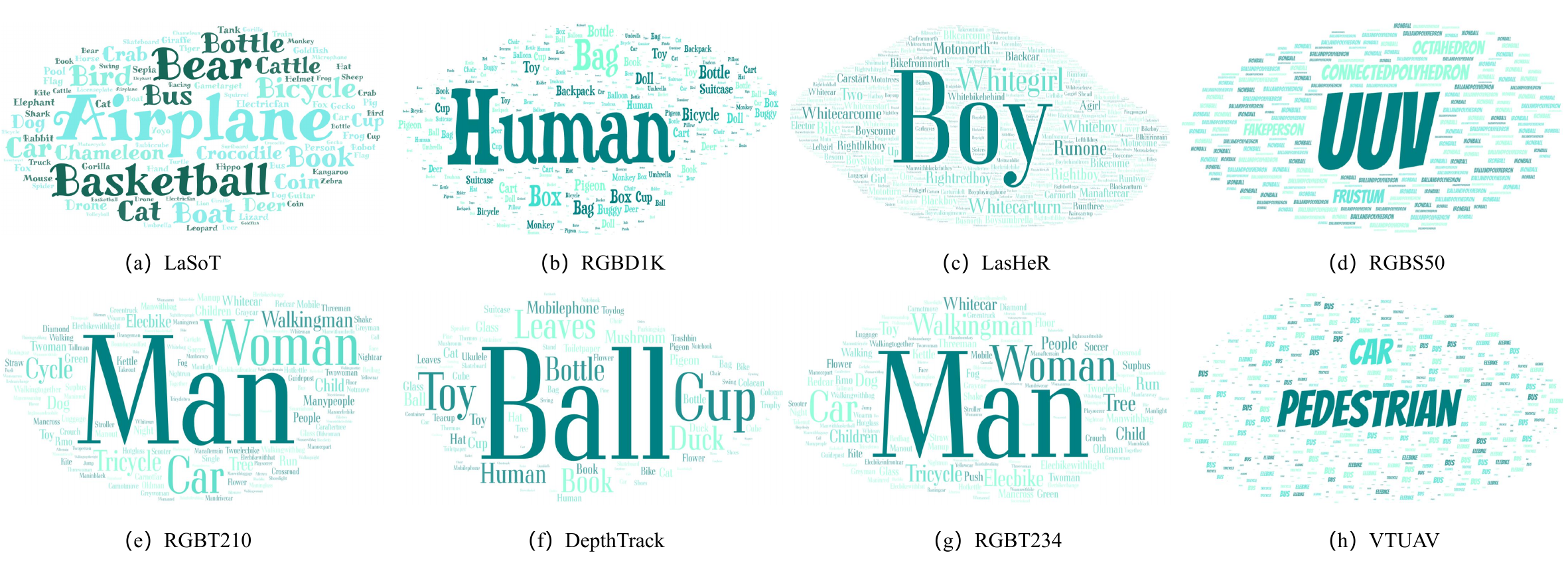}
\caption{The differences in the distributions of the RGB, RGB+D, RGB+T, and RGB+S datasets. Note, sequences relating to other MMVOT tasks are not named according to specific object categories and are therefore unsuitable for this visualisation, except for RGB+L datasets, which share the same sequences as the RGB datasets.
}
\label{fig:datadistribution}
\end{figure*}

Due to page limitations, Fig.~\ref{fig:resultsonbenchmarks} includes benchmarking results of four representative benchmarks, each displaying the performance of only the top 15 tracking methods.
These benchmarks are LasHeR \cite{lasher} (RGB+T), DepthTrack \cite{depthtrack} (RGB+D), VisEvent \cite{visevent} (RGB+E), and TNL2K \cite{tnl2k} (RGB+L).
As to the other two newly-emerged tasks, RGB+NIR and RGB+S, the maturity of research is very limited, which constrains in-depth discussion.
Additional details, including the performance of all methods across all benchmarks and tracking efficiency, are available on the project \href{https://github.com/Zhangyong-Tang/Awesome-MultiModal-Visual-Object-Tracking}{homepage}.

Specially, all unified trackers are summarised in Table.~\ref{tab:unified}, providing a direct and comprehensive comparison of algorithms targeting this emerging research area.
Among these methods, SUTrack \cite{sutrack} achieves leading performance across all evaluated benchmarks, which can be attributed to four key factors: (1) a more powerful backbone, (2) task token recognition, (3) learnable background/foreground enhancement, and (4) the inclusion of additional RGB data during training.
Among these, the utilisation of RGB data in the training phase is the most distinctive feature compared with other unified methods. 
This strategy is formulated as follows:
\begin{equation}
\begin{split}
        \rm T_{SUTrack} &:\rm (D_{RGB}, D_{MMVOT}, \theta_{RGB}) \to \mathbb{R}^4 \\
        \rm T_{Others} &:\rm (D_{MMVOT}, \theta_{RGB}) \to \mathbb{R}^4  \\
\end{split}
\label{eq:sutrack}
\end{equation}
where $\rm T_{SUTrack}$ and $\rm T_{Others}$ denote SUTrack and other unified trackers, respectively.
$\rm \theta_{RGB}$ represents a model pre-trained on RGB data.
As shown in Eqn.~\ref{eq:sutrack}, SUTrack achieves unification by fine-tuning the RGB pre-trained model, while also incorporating the original RGB data during training.
This strategy aligns with the concept of \textit{Replay} \cite{continuallearning2024} in the field of continual learning, which preserves the performance achieved on previous tasks, while acquiring knowledge for new ones.
By leveraging replay, SUTrack is able to retain the discriminative power of the RGB pre-trained model.
This is particularly beneficial for tracking, as RGB-only datasets typically offer higher-quality data than most MMVOT datasets.

\subsection{Challenging Attributes}
To thoroughly analyse the strengths and limitations of various methods, all datasets listed in Table.\ref{tab:dataset} are annotated with multiple challenging attributes.
In total, 61 distinct attributes are identified and categorised into task-shared and task-specific groups, based on their presence across tasks.
However, limited by pages, detailed descriptions of these attributes are provided in the supplementary material.

\section{Discussions}\label{sec:discussion}
\textbf{Discussion-A: Should multi-modal fusion be applied routinely or selectively?} 
Current research largely focuses on designing increasingly complex fusion models to boost benchmark performance, often overlooking a more fundamental question, the benefit of fusion in particular circumstances itself.
As illustrated in Fig.~\ref{fig:necessaryoffusion2}(a.1), most existing methods apply fusion in a dense manner, operating under the assumption that both modalities consistently offer high-quality information beneficial for accurate object localisation.
However, this assumption can be violated in both theoretical and practical contexts.
Theoretically, task definitions like RGB+T tracking are motivated by specific scenarios such as nighttime, where the RGB modality suffers from low visibility and the T data becomes crucial \cite{lasher}. 
We refer to such contexts as task-specific scenarios.
Practically, as shown in Fig.~\ref{fig:necessaryoffusion2}(b), low-quality inputs from one or more modalities can be observed in datasets across a range of tasks, including RGB+E, RGB+L, RGB+D, and RGB+T tracking. 
This confirms that degraded modalities are not uncommon in real-world data.
These theoretical motivations and empirical findings raise a critical question: Should multi-modal fusion be applied discriminately, such that low-quality modalities are excluded from the fusion process?
A conceptual illustration of this discriminate fusion paradigm is shown in Fig.~\ref{fig:necessaryoffusion2}(a.2), where fusion is selectively activated based on data quality.

To provide a more intuitive understanding, the right part of Fig.~\ref{fig:necessaryoffusion2} presents illustrative examples from RGB+T tracking.
First, histograms depicting data distributions from both common and task-specific scenarios are shown, where one modality exhibits insufficient quality to support tracking effectively, highlighting the importance of taking the variability of data quality across different conditions into consideration.
The subsequent quantitative analyses compare the performance of using RGB-only versus T-only inputs. 
The results indicate that using the less informative modality can yield a performance comparable to random noise (RGB-only vs. T-only: 0.732 vs. 0.002 and 0.006 vs. 0.378 on two cases). 
This finding suggests that fusion may be unhelpful or even harmful when one modality lacks any informative content.
Subsequently, the visualisations of the tracking results using RGB, T, and RGB+T (fused) inputs demonstrate that fusion does not always lead to improvements compared with the deployment of a single high-quality modality.
To sum up, these detailed examples strongly support the argument that multi-modal fusion should be applied discriminately. Routine fusion may inadvertently incorporate misleading or noisy information from the weaker modality, thereby degrading the overall tracking performance, especially in task-specific scenarios where the quality of the respective modalities is highly imbalanced \cite{mv-rgbt}.

\textit{Difficulty:} To address this, a data quality dependent fusion block must demonstrate robust performance across diverse scenarios. 
However, in some task-specific scenarios, no corresponding training datasets currently exist for all the tasks.  For instance, only a test set is provided for RGB+T tracking \cite{mv-rgbt}. This precludes the training of a discriminate fusion block.

\textbf{Discussion-B: Are existing multi-modal datasets capable of enabling sufficient generalisation?} 
In practical applications, the categories of tracked objects can be highly diverse, such as persons, vehicles, bags, and animals, depending on the scenarios.
This diversity imposes a requirement on the MMVOT training datasets to cover a wide range of object categories in order to enhance the generalisation capability.
To investigate this, Fig.~\ref{fig:datadistribution} presents a qualitative analysis of several MMVOT datasets alongside LaSOT \cite{lasot}, a representative RGB-only tracking benchmark. 
The figure visualises the distribution and frequency of object categories within each dataset, offering insights into their potential for achieving good generalisation.

Compared with LaSOT, it is evident that object categories in existing MMVOT datasets are highly concentrated, often exhibiting pronounced long-tail distributions.
Notably, we are not claiming that long-tail distributions are inherently undesirable.
In fact, they are more reflective of the natural data distributions \cite{vasttrack}.
However, such imbalance can lead to biased model capabilities.
In many cases, a MMVOT dataset contains a single dominant category. The models trained on such a dataset would perform significantly better on that category than on others.
This contradicts the fundamental objective of visual object tracking: to robustly follow any specified object, regardless of its category, throughout a sequence.

In addition, a closer inspection reveals that animals are significantly under-represented in current MMVOT datasets, with nearly no presence in Fig.~\ref{fig:datadistribution}.
This is noteworthy because animal tracking constitutes a common and practical application scenario, where multi-modal data is often leveraged to enhance robustness \cite{zoo145}.
The lack of multi-modal videos featuring animals therefore hinders progress in developing and evaluating trackers for such applications.
In summary, existing multi-modal datasets exhibit suboptimal object category distributions, with the notable absence of animal data posing a particular limitation.

\section{Future Work}
The preceding systematic review has comprehensively characterised the key components of MMVOT development, including multi-modal data collection (Sec.\ref{sec:data}), data alignment and annotation (Sec.\ref{sec:anno}), model development (Sec.\ref{sec:modeldesigning}), and evaluation methodology (Sec.\ref{sec:evaluation}), thereby establishing a holistic overview of the MMVOT field.
Building upon these, we identify open challenges and research issues within each section, offering valuable suggestions for future investigations.

\subsection{Multi-Modal Data Collection}
$\bullet$ \textit{Collecting Datasets with More Modalities:} 
Building unified trackers is currently a hot topic in the community. 
The models are typically trained on a mixture of multi-modal data from various tasks but are evaluated on benchmarks, each relating to a single type of task-specific data, leading to an inconsistency between training and testing procedures \cite{m3track, uastrack, sdstrack, vipt}.
To address this discrepancy, it would be valuable to construct a benchmark that incorporates multiple data modalities within a unified test set, thereby enabling more consistent and comprehensive evaluations of unified trackers.

$\bullet$ \textit{Collecting Datasets in Task-Specific Scenarios:} 
For convenience, existing datasets are mainly collected in common scenarios, like sunny days, where both modalities present sufficient quality for supporting tracking tasks. 
However, the emergence of MMVOT tasks is typically driven by real-world scenarios where RGB data quality is compromised \cite{mv-rgbt, visevent, rgbd1k, tnl2k, cmotb-aaai2022, rgbs50}.
This highlights a fundamental mismatch: current datasets often fail to represent the very conditions that motivate the use of multi-modal tracking methods.
Therefore, it is imperative to construct new datasets captured in task-specific scenarios that, by design, address the data gap between controlled research settings and practical applications.

$\bullet$ \textit{Collecting More Animal Data for MMVOT Tasks:} 
As discussed in Sec.~\ref{sec:discussion}, animal tracking in the wild represents a typical and meaningful application scenario, where the cooperation of multiple modalities is crucial for robust perception. 
However, this scenario is severely under-represented in existing MMVOT datasets, with a noticeable scarcity of multi-modal animal videos. 
To foster progress in this important direction, it is essential to collect and curate more multi-modal datasets specifically targeting animal tracking in diverse and challenging environments.

\subsection{Multi-Modal Data Annotation}
$\bullet$ MMVOT tasks are generally regarded as downstream extensions of the traditional VOT task.
Consequently, many configurations, including annotation protocols, are directly inherited. 
However, this practice does not always align with the physical characteristics of the additional modalities.
For instance, T data reflects endogenous thermal radiation, where object temperature gradually decreases from the interior to the exterior until reaching equilibrium with the environment.
This leads to inherently blurred object boundaries in thermal imagery, introducing uncertainty and ambiguity in bounding box annotations.
Therefore, it is pertinent to explore modality-specific annotation strategies that better reflect the characteristics of each modality, aiming to leverage their complementary strengths, while mitigating the impact of noisy or ambiguous cues.

 \subsection{Multi-Modal Model Design}
$\bullet$ \textit{Discriminate Fusion Paradigm:} 
The need for multi-modal fusion is often justified by task-specific scenarios. 
For instance, under extreme illumination conditions, RGB images may become unusable, while T data remains unaffected. However, the prevailing fusion paradigm in MMVOT applies dense fusion at every frame, regardless of the quality of each modality.
This observation highlights the limitations of indiscriminate fusion and underscore the importance of developing more selective, data quality dependent fusion paradigms, which can autonomously decide when and how to fuse modalities, improving robustness and reliability across both typical and adverse scenarios.

$\bullet$ \textit{Physics-Induced Architectures:} 
As introduced in Fig.~\ref{sec:datamodalities}, different modalities exhibit diverse physical characteristics.
However, due to the predominance of RGB pre-trained models, most auxiliary modalities (except L) are processed using similar operations, such as converting data into a 3-channel format to fit the input requirements of RGB models, thereby overlooking their unique physical properties.
To fully exploit the potential of these auxiliary modalities, it would be worth investigating the merits of developing physics-informed neural networks \cite{pinn} that explicitly incorporate their inherent physical characteristics, enhancing tracking performance, efficiency, and interpretability simultaneously.

$\bullet$ \textit{Multi-Modal LLMs in MMVOT:} 
Despite the significant breakthroughs in multi-modal LLMs \cite{mllms}, their application within the MMVOT domain is mainly limited to RGB+L tracking, remaining underexplored to a large extend. Given their broader knowledge topologies and reasoning capabilities, future research on integrating multi-modal LLMs into MMVOT tasks holds great potential, paving the way toward more reliable and general artificial intelligence in smart cities.

\subsection{Multi-Model Method Evaluation}
$\bullet$ Existing metrics for MMVOT tasks focus on evaluating tracking performance. 
Although this is the ultimate goal of tracking tasks, the assessment of multi-modal information fusion is largely overlooked and remains a black art. 
Many advanced methods claim improvements by more effective aggregation of multi-modal data. 
However, the improved performance only indicates that multi-modal information is better-used rather than better-fused. 
As illustrated in the right part of Fig.~\ref{fig:necessaryoffusion2}, superior results can be achieved by relying on a single modality rather than their fusion.
Therefore, developing metrics, which reflect the relationship between fusion and tracking performance better, is urgently required.  Measuring the quality and effectiveness of multi-modal fusion would offer a valuable insight that could be harnessed to improve tracking. 

\section{Conclusion}
\label{conclusion}

This work presents the first omni-survey on multi-modal visual object tracking (MMVOT) from the perspective of multi-modal data properties and analysis.
Our comprehensive discussion begins with an introduction to the physical characteristics of diverse modalities, revealing their complementary nature and providing a foundation for their deeper analysis.
Building on this, the survey, encompassing 338 papers, covers all aspects of constructing a model to perform a multi-modal tracking task. The topics include data collection, alignment, annotation, model architecture and training, as well as  evaluation methodology. 
Furthermore, we offer an in-depth discussion on two critical yet overlooked issues:
The first challenges the prevalent fusion paradigm, which assumes that multi-modal fusion is applied routinely in all circumstances. Instead, we advocate for selective, rather than indiscriminate, multi-modal information, depending on the quality of the data furnished by the respective modalities. This should avoid performance degradation caused by poor quality information.
The second highlights the extreme long-tail distribution in existing MMVOT datasets, which hampers generalisation. This is exemplified by the notable scarcity of animal data, limiting the advances that could be made specially in multi-modal animal tracking.
Finally, we identify and outline various open issues, providing valuable suggestions for future research.

In conclusion, this survey overviews the key elements of multi-modal visual object tracking and presents a deep analysis of all aspects of multimodality in MMVOT.
We hope it will serve as a valuable resource for researchers and practitioners, whether they are already familiar with MMVOT tasks or new to the field.
To support further advancement of the MMVOT community as well as its deployment in smart cities, the collated statistics, supplementary material, and the details of the reported analysis are publicly available at \textit{https://github.com/Zhangyong-Tang/MultiModal-Visual-Object-tracking} and will be continuously updated.

\bibliographystyle{IEEEtran}
\bibliography{ref.bib}

\vspace{-20pt}
\begin{IEEEbiographynophoto}{Zhangyong Tang}
is now a Ph.D. student with the School of Internet of Things Engineering, Jiangnan University. His research interests include multi-modal object tracking and information fusion. He has published several scientific papers, including IJCV, TOMM, CVPR, AAAI, ACMMM, etc. He achieved top 3 performance in several competitions, including the VOT2020 RGBT challenge (ECCV2020), VOT2021 D challenge (ICCV2021), VOT2022 RGBD challenge (ECCV2022), and the 2\textsuperscript{nd} Anti-UAV challenge.
\end{IEEEbiographynophoto}
\vspace{-10pt}
\begin{IEEEbiographynophoto}{Tianyang Xu}
(Member, IEEE) 
received his Ph.D. degree
at the School of Artificial Intelligence and Computer Science, Jiangnan University, Wuxi, China, in
2019. He is currently an Associate Professor at the
School of Artificial Intelligence and Computer Science, Jiangnan University, Wuxi, China. His research interests include visual tracking and deep learning.
He has published several scientific papers, including
IEEE TPAMI, IJCV, ICCV, IEEE TIP, IEEE TIFS,
IEEE TKDE, CVPR, ECCV, ICCV, etc. His publications have been cited more than 5000
times. He achieved top 1 performance in several competitions, including the
VOT2020 RGBT challenge (ECCV20), Anti-UAV challenge (CVPR20), and
MMVRAC challenges (ICCV2021, ICME2024).
\end{IEEEbiographynophoto}

\vspace{-20pt}
\begin{IEEEbiographynophoto}{Xuefeng Zhu}
received the Ph.D. degree in Jiangnan University, Wuxi, China, in 2024. He is currently a lecture with the School of Artificial Intelligence and Computer Science,
Jiangnan University. His research interests include visual tracking and deep
learning.
\end{IEEEbiographynophoto}
\vspace{-20pt}
\begin{IEEEbiographynophoto}{Hui Li}
received the Ph.D. degree in Jiangnan University, Wuxi, China, in 2021. He is currently an assistant professor with the School of Artificial Intelligence and Computer Science,
Jiangnan University. His research interests include image fusion and visual object tracking.
\end{IEEEbiographynophoto}
\vspace{-20pt}
\begin{IEEEbiographynophoto}{Shaochuan Zhao}
received the Ph.D. degree in Jiangnan University, Wuxi, China, in 2024. He is currently a post-doc with School of Computer Science and Technology, China University of Mining and Technology, Xuzhou 221116, China. His research interests include visual tracking.
\end{IEEEbiographynophoto}

\vspace{-20pt}
\begin{IEEEbiographynophoto}{Zhou Tao}
(Senior Member, IEEE) received the Ph.D. degree from Shanghai Jiao Tong University in 2016.
He is currently a Full Professor at the School of Artificial Intelligence and Computer Science, Jiangnan University, Wuxi, China.
His research interests include machine learning, computer vision, AI in healthcare, and medical image analysis. He is an Associate Editor of IEEE TNNLS, IEEE
TIP, IEEE TMI, and Pattern Recognition.
\end{IEEEbiographynophoto}
\vspace{-20pt}
\begin{IEEEbiographynophoto}{Xiao-Jun Wu}
received the Ph.D.
degree in pattern recognition and intelligent system
from the Nanjing University of Science and Technology, Nanjing, in 1996 and 2002, respectively. From
1996 to 2006, he taught at the School of Electronics
and Information, Jiangsu University of Science and
Technology, where he was promoted to Professor. He
has been with the School of Information Engineering, Jiangnan University, since 2006, where he is a
Professor of Computer Science and Technology. He was a Visiting Researcher
with the Centre for Vision, Speech, and Signal Processing, University of
Surrey, U.K., from 2003 to 2004. He has published over 300 papers. His
current research interests include pattern recognition, computer vision, and
computational intelligence. He was a Fellow of the International Institute for
Software Technology, United Nations University, from 1999 to 2000. He was
a recipient of the Most Outstanding Postgraduate Award from the Nanjing
University of Science and Technology.
He is currently a Professor in artificial intelligent and pattern recognition at
the Jiangnan University, Wuxi, China. He is IAPR/AAIA Fellow. His research
interests include pattern recognition, computer vision, fuzzy systems, neural
networks and intelligent systems.
\end{IEEEbiographynophoto}

\vspace{-20pt}
\begin{IEEEbiographynophoto}{Chunyang Cheng}
is currently a Ph.d student with the school of Artificial Intelligence and Computer Science, Jiangnan University. His research interests include image fusion.
\end{IEEEbiographynophoto}
\vspace{-20pt}
\begin{IEEEbiographynophoto}{Josef Kittler}
(Life Member, IEEE)
received the Ph.D. degree from the University of Cambridge, in 1974.
He is a distinguished Professor of Machine Intelligence at the Centre for Vision, Speech and Signal Processing, University of Surrey, Guildford, U.K.
He conducts research in biometrics, video and image database retrieval, medical image analysis, and cognitive vision. 
He published the textbook Pattern
Recognition: A Statistical Approach and over 700
scientific papers. His publications have been cited
around 70,000 times (Google Scholar).
He is series editor of Springer Lecture Notes on Computer Science. He
currently serves on the Editorial Boards of Pattern Recognition Letters, Pattern
Recognition and Artificial Intelligence, Pattern Analysis and Applications. He
also served as a member of the Editorial Board of IEEE Transactions on
Pattern Analysis and Machine Intelligence during 1982-1985. He served on
the Governing Board of the International Association for Pattern Recognition
(IAPR) as one of the two British representatives during the period 1982-2005,
President of the IAPR during 1994-1996.
\end{IEEEbiographynophoto}

\vfill

\end{document}


\title{Supplementary Material \\
Omni Survey for Multimodality Analysis \\ in Visual Object Tracking}

\author{Zhangyong Tang,
        Tianyang Xu~\IEEEmembership{Member,~IEEE,}
        Xuefeng Zhu,
        Hui Li,
        Shaochuan Zhao,
        Tao Zhou,
        Xiao-Jun Wu*,
        Chunyang Cheng,
        and Josef Kittler~\IEEEmembership{Life Member,~IEEE}
        \thanks{
        Z. Tang, T. Xu, X. Zhu, H. Li, X.-J. Wu, and C. Cheng are with the School of Artificial Intelligence and Computer Science, Jiangnan University, Wuxi 214122, China. (First author: Z. Tang: zhangyong\_tang\_jnu@163.com, Corresponding author: X.-J. Wu, e-mail: wu\_xiaojun@jiangnan.edu.cn)
        }
        \thanks{
        S. Zhao is with the School of Computer Science and Technology, China University of Mining and Technology, Xuzhou 221116, China. (e-mail: shaochuan\_zhao@cumt.edu.cn)
        }
        \thanks{
        T. Zhou is with the School of Computer Science and Engineering, Nanjing University of Science and Technology, Nanjing, 210094, China.(e-mail: : taozhou.ai@gmail.com)
        }
        \thanks{Josef Kittler is with the Centre for Vision, Speech and Signal Processing, University of Surrey, Guildford, GU2 7XH, U.K. (e-mail: j.kittler@surrey.ac.uk)}
}

\markboth{Journal of \LaTeX\ Class Files,~Vol.~14, No.~8, August~2021}%
{Shell \MakeLowercase{\textit{et al.}}: A Sample Article Using IEEEtran.cls for IEEE Journals}


\maketitle





\section{Introduction}\label{sec:introduction}
This is the official supplementary material for the survey work entitled "Omni Survey for Multimodality Analysis in Visual Object Tracking", which is the \textbf{first comprehensive survey in multi-modal visual object tracking (MMVOT) tasks from the perspective of multimodality analysis}.
Thus, limited by pages, detailed introductions of some components that are not highly related to multimodality are included in this supplementary file. 
They are:
\begin{itemize}
 	\item \textbf{A.} information of datasets.
        \item \textbf{B.} challenging attributes.
        \item \textbf{C.} evaluation metrics.
        \item \textbf{D.} loss function.
\end{itemize}

\vspace{5mm}
\textbf{A. Information of Datasets}

\textit{RGB+T}: OTCBVS \cite{otcbvs} and LITIV \cite{litiv} are two pioneering datasets, bringing the tracking task from RGB to RGB+T.
Despite their essential values, they are not the ones who make this task a hot topic due to their limited size, 6 and 9 videos\footnote{In this work, 'videos' represents multi-modal video pairs in MMVOT and videos in VOT.}, respectively. 
As the first RGB+T dataset with a certain scale, the proposal of GTOT \cite{gtot} is a flashpoint.
After that, contributed by the same group\footnote{https://chenglongli.cn/code-dataset/}, two larger datasets, RGBT210 \cite{rgbt210} and its extension RGBT234 \cite{rgbt234}, are published one after another, significantly accelerating RGB+T tracking.
For example, for the first time, RGB+T tracking is involved in the most well-known challenge in the tracking community, VOT Challenge\footnote{https://www.votchallenge.net/}.
This challenge evaluates all the videos in RGBT234 and selects a hard subset with 60 videos for the competitions in 2019 and 2020, named VOT-RGBT2019 \cite{votrgbt2019} and VOT-RGBT2020 \cite{vot-rgbt2020}, which have the same data but different protocols for evaluation.
However, compared to the training data size used in RGB tracking, there still exists an evident deficiency in data size, inducing the efforts for designing T-data synthesisers in data-driven \cite{eco-tir} and hand-crafted \cite{siamcda} ways.
Although the fake data leads to improvements in performance \cite{mfdimp, siamcda}, it can be easily distinguished from real T data and thus, most of the researchers still focus on the collection of real data.
In 2021, the first large-scale dataset, LasHeR \cite{lasher}, is published.
It has 1224 videos in total, 245 of them contained in the test set and 979 of them form the training set.
Later, another large-scale dataset, VTUAV \cite{vtuav}, is proposed with its data captured in the perspective of unmanned aerial vehicles (UAVs).
It contains 500 videos with each of them has an average length of 3329 frames.
Differently, \cite{mv-rgbt} revisits the current benchmarks and perceives that they are not representative for imaging conditions which motivate RGBT tracking since almost all of them are collected under common scenarios.
In response, the first benchmark highlighting the modality validity is proposed, termed MV-RGBT.

\textit{RGB+D}: The first RGB-D dataset, BoBoT-D \cite{bobotd}, is proposed in 2010, including 5 videos about \textit{milk}, \textit{tank}, \textit{ball}, \textit{person}, and \textit{lunch box}.
However, it is currently unavailable, which makes it not discussed in recent works.
Generally, PTB \cite{ptb} is widely believed as the foundation of RGB+D tracking.
It contains 100 videos collected indoor and thus, falls short in diversity.
Based on this, STC \cite{stc} is proposed as a diverse dataset by capturing data both in indoor and low-light outdoor scenarios.
Nevertheless, the scale of STC is smaller, with only 35 videos, which apparently leads the way for future work.
Consequently, CDTB \cite{cdtb} mitigates both the drawbacks of PTB and STC.
It captures data in three acquisition setups: (1) RGB+D sensors; (2) Time-of-flight (TOF)-RGB pairs; (3) Setero-RGB pairs.
This makes CDTB less limited when collecting outdoor videos, ensuring its diversity.
Totally, it has 80 videos and approximately 102K frames, 5x times about STC.
Accordingly, CDTB importantly supports the competitions held in the VOT Challenge from 2019 to 2022 \cite{votrgbt2019, vot-rgbt2020, vot2021, vot-rgbd2022},
In VOT-RGBD2022, another participated dataset is DepthTrack \cite{depthtrack}.
It has 200 videos and the first training set in RGB+D tracking with 150 videos, which makes great sense in the deep learning era.
However, this training subset still exhibits obvious limitation in terms of the number of videos, motivating the proposal of RGBD1K \cite{rgbd1k} and ARKitTrack \cite{arkittrack} in 2023.
RGBD1K and ARKitTrack consist of 1050 (1000 for training) and 455 (405 for training) videos, respectively.
RGBD1K is collected through manually integrated platform while a portable phone, iphone 13pro, is utilised for the collection of ARKitTrack, highly contributing its diversity in object categories and scenes.
Besides, ARKitTrack also provides pixel-level annotations to benefit the visual object segmentation (VOS) task \cite{vot2023}.
As to $\rm D^2Cube$ \cite{d2cube}, it is a newly published dataset collected by UAVs, including 900 videos for training and 100 videos for test.

\textit{RGB+E}: Inspired by the low latency of E data in motion perception, EED \cite{eed} only contains 7 videos and 234 frames, which gives reason to its negligence. 
To enable further investigations on jointly employing RGB and E data, FE108 \cite{fe108} is proposed with 108 videos and 20.9k frames in total.
FE240hz \cite{afnet} and FE141 \cite{fe141} are two extensions of FE108 by collecting more challenging videos.
Moreover, the first training set (500 videos) is published as a subset of VisEvent \cite{visevent} (820 videos) in 2021. 
COESOT \cite{coesot} is published in 2022 with a larger and more diverse training set.
Specifically, COESOT is collected from 90 different object categories,  containing 1354 videos in total and 827 of them belonging to the training set.
However, aforementioned datasets only focus on the short-term evaluations, which partially reflects the practical requirements for deployment.
Hence, FELT \cite{felt} is proposed with considerations on long-term tracking and its average length reaches 2148, which is approximately 6x times of that of COESOT.
Another issue reported in \cite{crsot} is about the multi-modal alignment, which is caused by the fact that RGB and E data are obtained through different sensors.
In detail, these sensors have different resolutions and view fields, making it difficult to be spatially aligned.
Additionally, the low latency of event sensors means there exists numerical event clips during imaging an RGB image, leading to difficulties for temporal alignment.
Therefore, CRSOT \cite{crsot} is proposed to encourage the explorations with unaligned multi-modal data, with its training and test sets including 836 and 194 videos, respectively.

\textit{RGB+L}: OTB99-L \cite{otb99_l} is the first dataset combining both visual and linguistic descriptions of a specific target.
As an extension of OTB100 \cite{otb100}, it contains 100 RGB videos and one sentence for each video.
However, its linguistic descriptions are solely originated from the first frame, such as \textit{Man with blue shirt and backpack next to a tree}, lacking in temporal considerations.
Accordingly, TNL2K \cite{tnl2k} and LaSOT \cite{lasot} are proposed with much larger data size, 2000 and 1400 videos, respectively, and natural language descriptions derived from the entire video.
Moreover, compared to LaSOT, LaSOT\_EXT \cite{lasot_ext} includes 150 videos and 0.4M frames in extra, resulting in an extended dataset.
So far, the aforementioned datasets have common properties: (1) short-term evaluations; (2) one kind of linguistic descriptions; (3) limited number of categories.
Based on these, MGIT \cite{mgit} captures 150 videos with an extreme average length for each video, reaching 13333 frames.
Besides, it encompasses language descriptions derived in a multi-granular annotation strategy: action, activity, and story, which is demonstrated a better approach to characterise the video content.
VastTrack \cite{vasttrack} aims to overcome the thrid problem caused by the limited number of categories.
It owns 2115 object categories recognised in 50610 videos, making it much more diverse than other datasets.
After that, for the first time, WebUOT-1M \cite{webuot1m} introduces natural language into underwater tracking task, expanding its application areas.
However, the motivation of integrating language descriptions, human-machine interaction, is still conducted once in the first frame, which poses insufficient utilisation of the language modality.
Based on this, VLT-MI \cite{vlt-mi} is established for multi-round interactions through relabelling the four published datasets, hoping to deepen the integration of RGB and language clues.
TO be more diverse, DTVLT \cite{dtvlt} further endows existing datasets with four kinds of text descriptions, \ie initial concise, initial detailed, dense concise, and dense detailed.

\textit{RGB+NIR}: 
CMOTB is the only dataset for RGB+NIR tracking.
It has 1000 videos in total and each video has 735 frames in average.
Furthermore, it can be divided into two subsets according to the difficulty for robust tracking: an easy set and a hard set.
The easy set contains 644 videos \cite{cmotb-aaai2022} and the remained 356 videos form the hard set \cite{cmotb-tnnls2024}.
Generally, the differences between these two subsets include: (1) more challenging scenarios in hard set; (2) more frequent presence of modality switch in hard set and the modality delay caused limited adaptability of cross-modal sensors in modality switch. 

\textit{RGB+S}: RGBS50 \cite{rgbs50} is the first dataset that combines RGB and S data.
It is proposed to remedy the disadvantages of RGB data in underwater scenarios, considering the characteristics of sonar sensors.
As its name indicates, it has 50 videos with an average length of 874 frames.
Notably, although the introduction of S data brings better robustness for underwater tracking task, its imaging principles result in the spatially unaligned multi-modal data (as shown in Fig.~\ref{fig:datamodalities}), which leads to its peculiarities in MMVOT community and causes further difficulties for method designing. 

\textit{RGB+Multi}: Different from other datasets containing two data modalities, WebUAV-3M \cite{webuav3m} and UniMod1K \cite{unimod1k} fuse three modalities to force more robust tracking systems, thus noted as RGB+Multi.
Specifically, RGB, L and audio data are involved in WebUAV-3M.
It collects 4500 videos from YouTube and endows a sentence description to each video, from which the audio annotations are generated through a text-to-speech software\footnote{http://balabolka.site/balabolka.htm}.
As to UniMod1K, it contains RGB, D, and L data simultaneously and is an extension of RGBD1K by endowing each video a language description. 
In conclusion, WebUAV-3M and UniMod1K are seeking to achieve robust trackers by means of combining more modalities.
Although it draws limited attention currently, great potential has already been demonstrated  in \cite{webuav3m, unimod1k}.

\vspace{5mm}
\textbf{B. Challenging Attributes}

\begin{table*}[h]
\footnotesize
\centering
\caption{Challenging attributes for dedicated analysis.}
\begin{tabular}{ccc}
\toprule
Category & Challenge & Description\\
\toprule
\multicolumn{1}{c}{\multirow{24}{*}{Shared (24)}}& OCC& Occlusion - the target is partially or fully occluded\\
         & NO& No occlusion - the target is not occluded\\
         & PO& Partially occlusion - the target is partially occluded\\
         & TO& Totally occlusion - the target is totally occluded\\
         & HO& Heavy occlusion - the target is heavily occluded (over 80\%)\\
         & LSV/SV& Large scale variation - the ratio of the first \\
         & & bounding box and the current bounding box is out of the range {[}0.5, 1{]} \\
         & LI& Low illumination - the illumination in the target region is low\\
         & HI& High illumination - the illumination in the target is too strong to identify the target\\
         & EI& Extreme illumination - the target is in low or high light condition\\
         & SO& Small object - the number of pixels in the groundtruth bounding box is less than 400\\
         & DEF& Deformation - non-rigid object deformation\\
         & LR& Low resolution - the resolution in the target region is low\\
         & BC& Background clutter - the background information which included the target is messy\\
         & MB& Motion Blur - the target motion results in the blur image information\\
         & CM& Camera moving - the target is captured by moving camera\\
         & TB& Target blur - target is blurry caused by illumination ot motion\\
         & FM& Fast Motion - the motion of the groundtruth is larger than 10 pixels\\
         & SA/ST& Similar appearance - there are objects of similar appearance near the target\\
         & ARC/AC& Aspect ratio change - the ratio of bounding box aspect is outside the range {[}0.5, 2{]}\\
         & OV& Out of view - the target is completely missing in the current view\\
         & OP& Out-of-plane rotation - the target suffers out-of-plane rotation\\
         & IP& In-plane rotation - the target suffers in-plane rotation\\
         & ROT& Rotation - target rotates\\
         & VC& Viewpoint change - the viewpoint is not fixed because the capturing angle changes\\
\midrule
\multicolumn{1}{c}{\multirow{5}{*}{RGB+T (5)}}& TC& Thermal crossover - the target has similar temperature with other objects or background\\
         & HO& Hyaline occlusion - the target is occluded by hyaline object (LasHeR)\\
         & AIV& Abrupt illumination variation - the illumination of the target changes significantly\\
         & FL& Frame lost - some of the thermal frames are lost\\
         & TVS& Thermal-visible separation - the bounding boxes in visible and thermal images have no overlap\\
\midrule
\multicolumn{1}{c}{\multirow{12}{*}{RGB+D  (12)}}& DV& Depth variation - depth change of all pixels inside the bounding box (relatively ratio)\\
         & CDV& Colour distribution variation - RGB distribution change of the bounding box\\
         & DDV& Depth distribution variation - depth distribution change of the bounding box\\
         & SDC& Surrounding depth clutter - depth similarity between the \\
         &&target and ring-shaped contextual region (mean value)\\
         & BSC& Background shape camouflages - the object in the \\
         &&background shares the similar shape as the target (binary value)\\
         & DS& Dark scene - the target is in dark scene\\
         & RT& Reflective target - the target is reflective\\
         & IC& Illumination change - there are illumination changes during one video sequence\\
         & CO& Composite objects - the target is an ensemble of multiple objects\\
         & NaN& Unassigned - no attributes are assigned\\
         & SF& Sensor failure - at least one camera can not provide useful information\\
         & LD& Low depth quality - the quality of depth information is low\\
\midrule
\multicolumn{1}{c}{\multirow{5}{*}{RGB+E (5)}}& MT& Moving targets - there exist 1 to 3 moving targets\\
         & WiB& What is the background - the background occupies the space between the camera and the target\\
         & HDR& High dynamic range\\
         & NMO& No motion\\
         & BOM& Background object motion - the motion of target is affected \\
         &&by the motion of background objects\\
\midrule
\multicolumn{1}{c}{\multirow{9}{*}{RGB+L (9)}}& AS& Adversarial samples - influence of adversarial samples\\
         & INV& Invisible - target is partially or fully invisible due to occlusion or out-of-view\\
         & PTI& Patial target information\\
         & NAO& Natural or artificial object\\
         & CAM& Camouflage targets\\
         & UV& Underwater visibility\\
         & WCV& The colour of water of the target area\\
         & US& Different underwater scenarios\\
         & LEN& The length of videos\\
\midrule
\multicolumn{1}{c}{\multirow{2}{*}{RGB+S (2)}}& LSR& Low sonar reflection - the target has a low luminance value in grayscale sonar image\\
         & SC& Sonar crossover - the target has similar sonar reflection value with surrounding objects\\
\midrule
\multicolumn{1}{c}{\multirow{3}{*}{RGB+NIR (3)}}& MA& Modality adaption - slow modality switch with pseudo intermediate modality\\
         & MM& Modality mutation - rapid modality switch without pseudo intermediate modality\\
         & MD& Modality Delay - Delayed modality switch due to inadequate adaptation\\
    \midrule
\multicolumn{1}{c}{\multirow{1}{*}{RGB+Multi (1)}}& COM& Complexity of the current video \\          
\bottomrule
\end{tabular}
\label{tab:attributes}
\end{table*}

To thoroughly analyse the pros and cons of methods, almost all the datasets are annotated with several challenging attributes.
As displayed in Table.~\ref{tab:attributes}, there are 61 different attributes in total. They are categorised as task-shared and task-specific ones according to their presence in different tasks, commonly-appeared attributes are thought task-shared ones and vice versa.

Task-shared: This category mainly includes attributes describing the target itself and the process of data collection, which can appear in all MMVOT tasks, such as occlusion (OCC), scale variation (SV), aspect ratio change (ARC), out-of-plane rotation (OP), deformation (DEF), small objects (SO), low resolution (LR), out-of-view (OV), camera motion (CM), and view change (VC). 
Besides, since all MMVOT tasks share the same data modality-RGB, attributes can be distinctly recognised in RGB images, like background clutter (BC), similar targets (ST), extreme illumination (EI), fast motion (FM), and motion blur (MB), belong to this category. 

Task-specific: There are several attributes highly related to the specific task or only appear in specific datasets.
For example, thermal-related ones, such as thermal crossover (TC), frame lost (FL), and thermal-visible separation (TVS), only appear in RGB+T datasets.
Analogously, depth- and distribution-related ones, including depth variation (DV), depth distribution variation (DDV), and low depth quality (LD)), belong to RGB+D tracking because depth information is only involved in this task and is widely used to compute the depth consistency in the format of distribution.  
Besides, since the only dataset supporting the underwater RGB+L tracking, WebUOT-1M, attributes like underwater visibility (UV) and different underwater scenarios (US) are thought unique to RGB+L tracking. 
RGB+E tracking has several attributes specific for motion due to the characteristics of event data, like moving targets (MT), no motion (NMO), and background object motion (BOM).
RGB+S tracking has two sonar-specific attributes, low sonar reflection (LSR) and sonar crossover (SC).
As to RGB+NIR tracking, the unique attributes are all related to the modality switch process, which is the core difference between this task and other MMVOT tasks.
They are termed as modality adaption (MA), modality mutation (MM), and modality delay (MD).
In WebUAV-3M, a dataset with three modalities, the complexity of videos is also quantified, resulting in the attribute COM.

Through the detailed analysis on attributes, insightful conclusions can be drawn, offering intuitive suggestions for further model modifications. 

\vspace{5mm}
\textbf{C. Evaluation Metrics}

Since MMVOT tasks are highly related to the popular RGB-only tracking task, most of the MMVOT datasets share the same metrics as RGB tracking datasets do.
To be self-contained, all the widely used metrics, including precision rate (PR) \cite{lasher}, success rate (SR), normalised precision rate (NPR) \cite{trackingnet}, precision (Pr) \cite{rgbd1k}, recall (Re), F-score, and the VOT protocol \cite{vot-rgbt2020, votrgbt2019}, are introduced in detail in the following paragraphs.

In advance, we give the introductions of two fundamental metrics, Intersection over Union (IoU) and  Centre position error (CPE):
\begin{equation}\label{eq:ioucpe}
\begin{split}
 \rm IoU & = \frac{|\textbf{\textit{gb}} \cap \textbf{\textit{b}}|}{|\textbf{\textit{gb}} \cup \textbf{\textit{b}}|}\\
 \rm CPE & =\rm  \sqrt{(c_{x, gb} - c_{x, b})^2 + (c_{y, gb} - c_{y, b})^2}
\end{split}
\end{equation}
where $\cap$ is an operation for computing the overlapped area while $\cup$ measures the total area covered by regions.
$|\cdot|$ indicates the number of pixels in the region.
$\textbf{\textit{gb}}$ and $\textbf{\textit{b}}$ denote the groundtruth and predicted bounding boxes, respectively.
Generally, they are represented by the coordinates of the top-left point, width, and height ($\rm \ell_x$, $\ell_y$, $\rm w$, $\rm h$).
Subscripts $x$ and $y$ are the notations of x- and y-axis.
$c_{x, gb}$ is the x coordinate of the centre point of $\textbf{\textit{gb}}$, and others are termed in the same way.

PR and NPR: Although averaging the CPE of all frames provides a measurement of accuracy, it fails to be correct when encountering tracking failures \cite{otb100}, which is because that the averaged CPE can be biased by several extremely bad predictions.
Based on these observations, PR and NPR are proposed as formulated:
\begin{equation}\label{eq:prnpr}
\begin{split}
\rm PR & = \frac{1}{\textit{N}}\sum_{\textit{i}=1}^{\textit{N}} \textit{pr}_\textit{i},\ \rm where \ \textit{pr}_\textit{i} = \left\{
\begin{aligned}
    & 1, \ \rm if \ \rm CPE_\textit{i} < \tau_{PR}, \\
    & 0, \ \rm if \ \rm CPE_\textit{i} >= \tau_{PR}. \\
\end{aligned}
\right. \\
\rm NPR & = \frac{1}{\textit{N}}\sum_{\textit{i}=1}^{\textit{N}} \textit{npr}_\textit{i},\ \rm where \ \textit{npr}_\textit{i} = \left\{
\begin{aligned}
     & 1, \ \rm if \ \rm NCPE_\textit{i} < \tau_{NPR}, \\  
    & 0, \ \rm if \ \rm NCPE_\textit{i} >= \tau_{NPR}. \\
\end{aligned}
\right.
\end{split}
\end{equation}

where $N$ is the number of frames.
$pr_i$ indicates whether prediction for the $i$-th frame is accurate or not, $pr_i = 1$ if the corresponding $\rm CPE_\textit{i}$ below the specific threshold $\tau_{PR}$ and vise versa.
Eventually, $\rm PR$ represents the percentage of accurately tracked frames.
However, $\rm PR$ is sensitive to the size of object and image resolution, leading to unfair evaluations for objects with different scales \cite{trackingnet}.
Hence, $\rm NPR$ is proposed by normalising CPE with the width and height of groundtruth bounding box $\rm \textbf{\textit{gb}}$.
This means $\rm NCPE=\sqrt{(c_{x, gb}/w - c_{x, b}/w)^2 + (c_{y, gb}/h - c_{y, b}/h)^2}$.
Except the normalisation, all other parameters in $\rm PR$ and $\rm NPR$ share the same meanings but different notations. 

SR: As Eq.~\ref{eq:sr} displays, $\rm SR$ is a metric defined in a way similar to PR.
It measures the robustness of trackers and is formulated as:
\begin{equation}\label{eq:sr}
\begin{split}
\rm SR & = \frac{1}{\textit{N}}\sum_{\textit{i}=1}^{\textit{N}} \textit{sr}_\textit{i},\ \rm where \ \textit{sr}_\textit{i} = \left\{
\begin{aligned}
    & 1, \ \rm if \ \rm IoU_\textit{i} >= \tau_{SR}, \\
    & 0, \ \rm if \ \rm IoU_\textit{i} < \tau_{SR}. \\
\end{aligned}
\right. 
\end{split}
\end{equation}
where $sr_i$ indicates whether the $i$-th frame is successfully tracked or not, $sr_i = 1$ if the corresponding $\rm IoU_\textit{i}$ exceeds the specific threshold $\tau_{SR}$ and vise versa.
The overall SR is then obtained through averaging the evaluations for all $N$ frames.

Pr, Re, and F-score: They are long-term metrics borrowed from \cite{vot18} and formulated as:
\begin{equation}\label{eq:prref}
\begin{split}
\rm Pr & = \frac{1}{\textit{N}_{pr}}\sum_{\textit{i}=1}^{\textit{N}} \rm IoU_\textit{i}, \rm where\ \textbf{\textit{b}}_\textit{i} \neq \emptyset \\
\rm Re & = \frac{1}{\textit{N}_{re}}\sum_{\textit{i}=1}^{\textit{N}} \rm IoU_\textit{i}, \rm where\ \textbf{\textit{gb}}_\textit{i} \neq \emptyset \\
\rm F\text{-}score & = \rm \frac{2Pr*Re}{Pr+Re} \\
\end{split}
\end{equation}
where $*$ represents multiplication.
$N_{pr}$ and $N_{re}$ denote the number of frames where the prediction and groundtruth are not empty, respectively.
Based on these, Pr and Re reflect different aspects of the methods, and thus, F-score is proposed for a comprehensive evaluation. 

VOT protocol: The committee of VOT challenge \cite{vot15, vot-rgbt2020} employs accuracy (A), robustness (R), and expected average overlap (EAO) for evaluation. 
Generally, A is computed in a way similar to Pr, averaging the IoU scores for all the valid frames.
R reflects the number of tracking failures, indicated by IoU = 0 since the VOT evaluation is implemented in a supervised approach.
Furthermore, EAO is proposed as a comprehensive metric, combining A and R and providing a more reasonable reference for ranking.
As its name suggests, it is also a metric similar to Pr but calculated within segments (Before 2020, the trackers will be re-initialised when a tracking failure is detected. After 2020, the VOT challenges use a multi-start strategy, which means trackers will not be online re-initialised any more and they will be evaluated from multiple pre-defined anchors with a fixed interval \cite{vot-rgbt2020}.
In both protocols, the interval between the initialisation point and the next failure is called a segment.)
Since different segments varies from the length, the segments shorter than a pre-defined length $N_{EAO}$ will be zero padded.
In this way, the final EAO can be obtained through averaging the IoU scores of all segments.

Notably, there are several metrics with thresholds, including PR, NPR, SR, and EAO.
Evidently, different thresholds will lead to unequal scores, inducing extra bias to performance measurement.
Hence, the performance of trackers are usually measured multiple times under different thresholds, mitigating the potential bias introduced by a specific threshold.
Connecting these performance scores in plots will form a curve and the area under curve (AUC) is believed reliable, acting as the final score.

Despite the successful application of above metrics, there are two pioneering works underscoring the significance of multi-modal specified evaluation strategies.
MV-RGBT \cite{mv-rgbt} proposes an RGB+T benchmark with all videos captured under extreme challenging scenarios.
Based on this, it is divided into two subsets, MV-RGBT-RGB and MV-RGBT-TIR, with the suffix RGB and TIR mean TIR and RGB data is corrupted, respectively.
Thus, a fine-grained compositional evaluation approach is established, which offers a quantitative perspective of modality balance within algorithms, providing in-depth suggestions for further modification.
VLTVerse \cite{vltverse} defines a relatively more complete evaluation space.
It contains 6 different kinds of linguistic annotations for each video, which can be actively determined by users, better showcasing the superiority of methods and also forming a multi-grained evaluation approach for a comprehensive analysis.

\vspace{5mm}
\textbf{D. Loss Function}

Usually, multi-modal information is already aggregated before the output stage.
This means there solely exists the loss function for one modality (could be RGB, TIR, or the fused modality).
In this way, the loss function for a multi-modal tracker is exactly the same with that utilised in RGB-only trackers \cite{ostrack}.

Generally, the tracking loss $\rm Loss$ includes two parts: classification $\rm Loss_{c}$ and regression $\rm Loss_{r}$. 
$\rm Loss_{c}$ describes the possibility of an image patch being the targeted patch.
Each patch is endowed a prior possibility using the cosine window based on the assumption that the object is more likely to be in the centre and has a predicted score.
Based on these, $\rm Loss_{c}$ is implemented by focal loss in advanced multi-modal trackers \cite{vipt, gmmt, rgbd1k}.
$\rm Loss_{r}$ computes the distance between the predicted bounding box and groundtruth bound box.
In \cite{vipt, gmmt, rgbd1k}, it consists of two metrics: $\rm Loss_{iou}$ constrains the overlap between two bounding boxes and $\rm Loss_{l1}$ shrinks the absolute distance of edges.
$\rm Loss_{iou}$ and $\rm Loss_{l1}$ are implemented by eq.\ref{eq:ioucpe} and $l1$-norm, respectively.
Therefore, 
\begin{equation}\label{eq:prref}
\begin{split}
\rm Loss & =\rm \alpha * Loss_c + \lambda* Loss_{iou} + \gamma* Loss_{l1}
\end{split}
\end{equation}

where $\alpha, \lambda,$ and $\gamma$ are set to 1, 2, and 5, respectively.

\bibliographystyle{IEEEtran}
\bibliography{ref.bib}

\vfill